\documentclass{article}

\usepackage{arxiv}

\usepackage[utf8]{inputenc} 
\usepackage[T1]{fontenc}    
\usepackage{amssymb}
\usepackage{amsmath}
\usepackage{bm}
\usepackage{url}            
\usepackage{booktabs}       
\usepackage{amsfonts}       
\usepackage{nicefrac}       
\usepackage{microtype}      
\usepackage{lipsum}		
\usepackage{graphicx}
\usepackage{subcaption}
\usepackage{natbib}
\usepackage{doi}

\title{ALBU: An approximate Loopy Belief message passing algorithm for LDA to improve performance on small data sets}

\author{\href{https://orcid.org/0000-0001-8646-3839}{\includegraphics[scale=0.06]{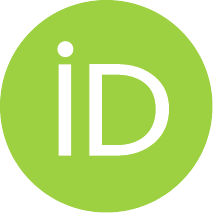}\hspace{1mm} Rebecca M.C.~Taylor} \\
	Department of Electrical and Electronic Engineering\\
	Stellenbosch University\\
	South Africa \\
	\texttt{becci.rr@gmail.com} \\

	\And
	\href{https://orcid.org/0000-0001-6775-9220}{\includegraphics[scale=0.06]{orcid.pdf}\hspace{1mm}Johan A. du Preez} \\
	Department of Electrical and Electronic Engineering\\
	Stellenbosch University\\
	South Africa\\
	\texttt{dupreez@sun.ac.za} \\
}

\date{}

\renewcommand{\shorttitle}{\textit{arXiv} Template}

\hypersetup{
pdftitle={A template for the arxiv style},
pdfsubject={q-bio.NC, q-bio.QM},
pdfauthor={David S.~Hippocampus, Elias D.~Striatum},
pdfkeywords={First keyword, Second keyword, More},
}

\begin{document}
\maketitle

\begin{abstract}
Variational Bayes (VB) applied to latent Dirichlet allocation (LDA) has become the most popular algorithm for aspect modeling. While sufficiently successful in text topic extraction from large corpora, VB is less successful in identifying aspects in the presence of limited data. We present a novel variational message passing algorithm as applied to Latent Dirichlet Allocation (LDA) and compare it with the gold standard VB and collapsed Gibbs sampling. In situations where marginalisation leads to non-conjugate messages, we use ideas from sampling to derive approximate update equations. In cases where conjugacy holds, Loopy Belief update (LBU) (also known as Lauritzen-Spiegelhalter) is used. Our algorithm, ALBU (approximate LBU), has strong similarities with Variational Message Passing (VMP) (which is the message passing variant of VB). To compare the performance of the algorithms in the presence of limited data, we use data sets consisting of tweets and news groups. Additionally, to perform more fine grained evaluations and comparisons, we use simulations that enable comparisons with the ground truth via Kullback-Leibler divergence (KLD). Using coherence measures for the text corpora and KLD with the simulations we show that ALBU learns latent distributions more accurately than does VB, especially for smaller data sets.
\end{abstract}


\section{Introduction}
\label{sec:Introduction}
LDA \cite{blei2003latent} is a hierarchical Bayesian model that is most commonly known for its ability to extract latent semantic topics from text corpora. It can also be thought of as the categorical/discrete analog of Bayesian principal component analysis (PCA) \cite{buntine2002variational} and can be used to extract latent aspects from collections of
discrete data \cite{blei2003latent}. It finds application in fields such as medicine \cite{backenroth2017fun, pritchard2000inference}, computer vision \cite{fei2005bayesian, sivic2005discovering, wang2008spatial} and finance \cite{feuerriegel2018investor, shirota2014extraction}. In non-text specific applications, it is commonly referred to as grade of
membership \cite{woodbury1982new} or aspect modeling \cite{minka2002expectation}.

In LDA the required posterior distribution is intractable (i.e., one cannot apply exact inference to calculate it) and therefore a variety of approximate inference techniques can be employed \cite{blei2003latent}. For most text topic modeling problems the traditional approximate inference techniques are sufficient. However, in many cases where less data is available (due to smaller collections and/or collections where document size is small) or where there is large topic overlap (these make for more difficult inference) alternative approaches are favored \cite{phan2008learning,hong2010empirical,ramage2010characterizing,jin2011transferring}. 

The phenomenal increase in social media usage, typified by digital microblogging platforms such as Twitter \cite{weng2010twitterrank, ramage2010characterizing}, has called attention to the need for addressing problems associated with the semantic analysis of text corpora involving short texts \cite{hong2010empirical, basave2014automatic, nugroho2015incorporating}. The difficulties inherent in the extraction of topic information from very short texts \cite{phan2008learning,mehrotra2013improving,yan2013biterm} are partly due to the small number of words remaining after the removal of stop words \cite{hong2010empirical,albakour2013sparsity}. On average, the number of words per tweet, even before data cleaning, is between 8 and 16 words \cite{jabeur2012uprising, celikyilmaz2010probabilistic}. 

One way of artificially increasing document length is to pool documents into longer pseudodocuments. This can be achieved by employing automatic strategies \cite{quan2015short}; by combining documents from the same author \cite{hong2010empirical, weng2010twitterrank}, by making use of temporal information \cite{mehrotra2013improving}, or by merging documents based on hashtags \cite{mehrotra2013improving,jin2011transferring}. These methods can be applied only if the relevant information is available and sufficient \cite{sokolova2016topic, ramage2010characterizing}.  Another approach to improve LDA performance in the presence of minimal data is to extend the graphical model itself to include additional information such as temporal \cite{wang2006topics} or author \cite{rosen2004author} information.

For small corpora (not necessarily consisting of documents containing short texts), a way to further improve performance is to apply transfer learning \cite{phan2008learning,hong2010empirical, jin2011transferring}; word-topic distributions are learnt from a large, universal corpus \cite{phan2008learning, jin2011transferring}, following which they are updated by applying LDA to the smaller corpus. However, due to limited general universal corpora for certain non-text applications as well as for niche context specific text corpora (legal and financial for example), results can be less than optimal.

In this article, we present our general approach to improving LDA accuracy on small data sets by using an alternative approximate inference technique. By modifying the inference only, we allow other strategies such as pooling or transfer learning to be applied additionally where the data set allows. 

\section{Variational inference}
Variational inference (VI) is an approximate inference technique that directly optimizes the accuracy of an approximate posterior distribution \cite{winn2004variational} that is parameterized by free variational parameters \cite{blei2017variational}. In VI, we choose a restricted family of distributions $q(\bm{h})$ (with $\bm{h}$ the variational parameters) and then find the member of this family (by finding the setting of the parameters) for which a divergence measure is minimized -- this changes an inference problem into an optimization problem \cite{blei2017variational}. 

Variational Bayes (VB), the original inference technique used in LDA \cite{blei2003latent}, is a framework that can be used to find $q(\bm{h})$ in an iterative expectation maximization like manner \cite{attias2000variational} 
once the variational equations have been derived. An alternative to deriving these equations for each model and then using VB for the optimization, is to use variational message passing (VMP) \cite{winn2004variational, winn2005variational}.  

The success of VB as an an approximate inference technique for LDA has resulted in its widespread use for topic modeling by machine learning practitioners \cite{foulds2013stochastic}, two of the main Python implementations being Gensim \cite{rehurek2010software} and Scikit-learn \cite{pedregosa2011scikit}. However, when working with small text corpora, or performing aspect modeling on small non-text data sets, the quality of extracted aspects can be low \cite{hong2010empirical, yan2013biterm}. In this article we do not alter the LDA graphical model, nor do we use document pooling or transfer learning, but instead we present method that employs the full joint distributions arising from the graphical description of the standard LDA model.

In our approach, we use a variational alternative to VB that is based on Loopy Belief update (LBU) \cite{koller2009probabilistic}[p364-366], also known as the Lauritzen-Spiegelhalter variant \cite{lauritzen1988local} of the Sum-Product algorithm \cite{heskes2003stable}. LBU is a message passing algorithm that has been shown to be variational in nature \cite{koller2009probabilistic} and differs from LBP in the manner that over-counting is handled (instead of excluding reverse direction messages when calculating an outgoing message, we perform message cancelling when updating the target factor. We utilize LBU directly where possible, and where approximate messages are required, we derive update equations by using a sampling approach (which turns out to have strong similarities with the VMP approximate messages).

ALBU is, to our knowledge, a novel novel algorithm for LDA since the only similar approach to Loopy Belief propagation (LBP) for LDA that we are aware of is a collapsed Loopy Belief implementation by Zeng \cite{zeng2012topic, zeng2012learning} where a Sum-Sum approximate of the Sum-Product algorithm \cite{zeng2012learning} is used. Due to the collapsed nature of the graphical model, the update equations differ significantly and the complexity of updating the Dirichlet distributions disappears. Other collapsed algorithms such as CVB \cite{teh2007collapsed} and CVBO \cite{asuncion2009smoothing} have shown to outperform VB but we do not attempt currently attempt to compare our work to these collapsed algorithms.

To compare our implementation with VB, we use an open source VB implementation, Gensim \cite{rehurek2010software}. We also benchmark our algorithm against collapsed Gibbs sampling to give an indication of its performance (using the standard Python lda package from PyPi \cite{pypi}) since is shown to be more effective than VB in extracting topics correctly \cite{teh2007collapsed, zeng2012learning}, especially in the presence of little data.

\section{The ALBU algorithm}
\label{sec:emdw}
In the following section we present the approximate LBU (ALBU) LDA algorithm. We first define the graph and then present the factor updates.
\subsection{Notation and graphical representation of LDA}

Because LDA was popularized in the context of text topic modeling \cite{blei2003latent}, the terminology for describing the algorithm is often linked to the terminology used in this field.  We are using simulated data and are therefore not bound by terminology from particular fields. For readability, however, we give preference to the traditional terminology used in text topic modeling: when applied to a corpus, LDA extracts topic-document Dirichlet distributions and word-topic Dirichlet distributions. 
\begin{figure}[ht]
\vskip 0.2in
\begin{center}
\centerline{\includegraphics[width=0.6\columnwidth]{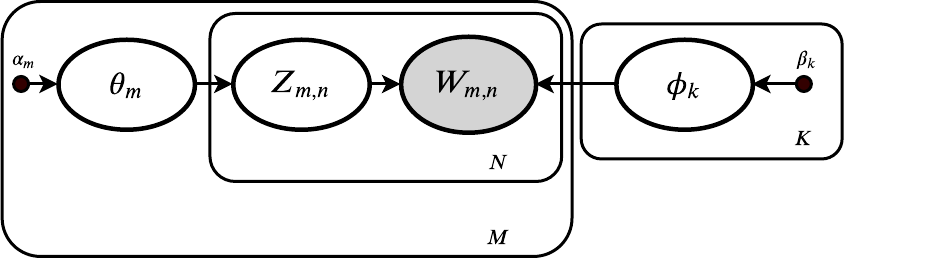}}
\caption{Plate model of LDA system as a Bayes net.\label{fig:bn}}
\end{center}
\vskip -0.2in
\end{figure}

\begin{figure}[ht]
\vskip 0.2in
\begin{center}
\centerline{\includegraphics[width=0.6\columnwidth]{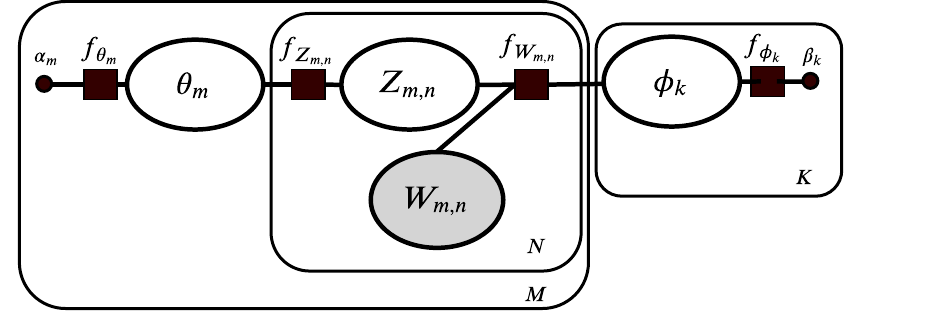}}
\caption{Plate model of LDA system as a factor graph. Variable nodes are given the name of their respective random variables and factor nodes are denoted by an $f$ and identified by the variable to the left of the conditioning bar on initialization, for example $f_{\theta_m} = p(\bm{\theta}_{m}|\bm{\alpha}_{m})$.\label{fig:bncg}}
\end{center}
\vskip -0.2in
\end{figure}
LDA was originally depicted \cite{blei2003latent} in the Bayes net plate model format shown in Fig.~\ref{fig:bn}. The $n$'th word in document $m$ is $W_{m,n}$ and 
depends on the (latent) topic $Z_{m,n}$ present in the document, which, in its turn selects a distribution from the set of Dirichlets  $\bm{\Phi} \equiv \{\bm{\phi}_1\ldots \bm{\phi}_{K}\}$ which describes the words present in each topic. 
\begin{figure}[ht]
\vskip 0.2in
\begin{center}
\centerline{\includegraphics[width=0.7\columnwidth]{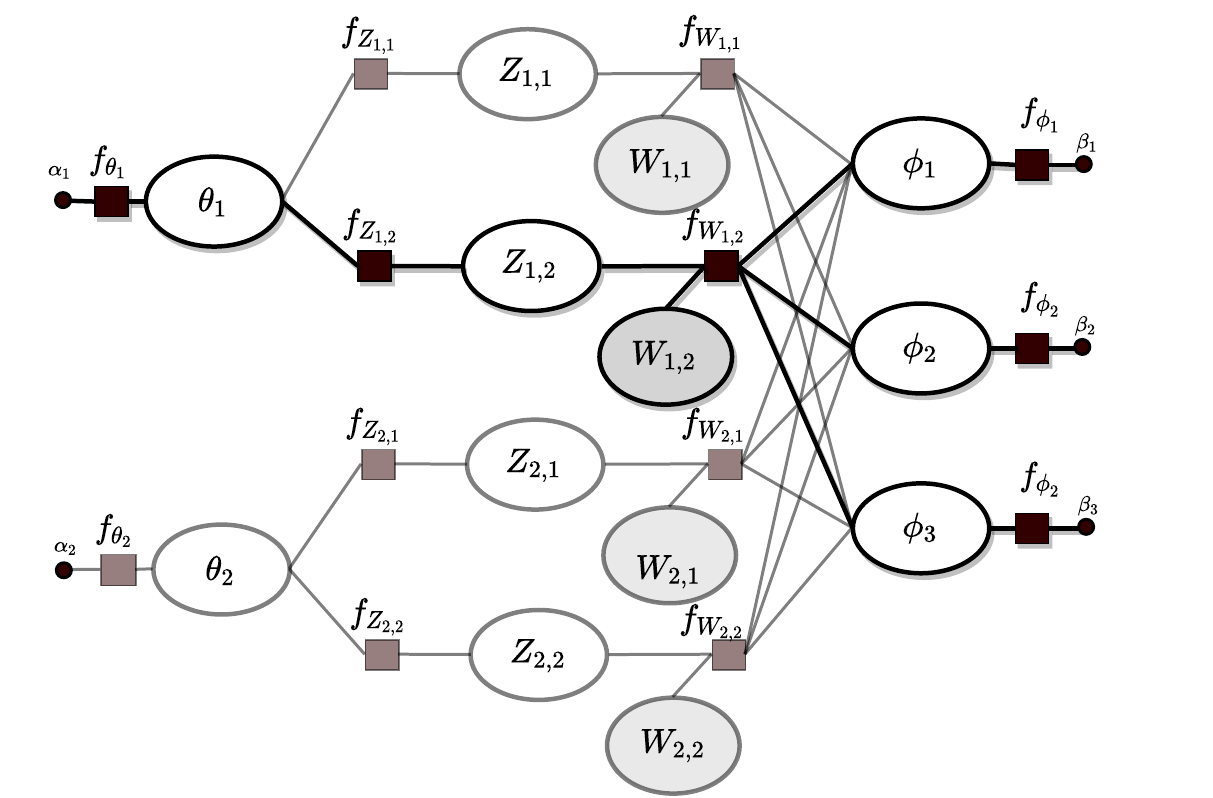}}
\caption{Unrolled factor graph representation of LDA for a two document corpus with two words per document. A single \textit{branch} is highlighted (where $m =1$ and $n = 2$). \label{fig:LDA_branch_hilight} }
\end{center}
\vskip -0.2in
\end{figure}

In Fig.~\ref{fig:bncg}, we show the factor graph representation of LDA where factors are named based on the random variables which are initially to the left of the conditioning bar. Figure \ref{fig:LDA_branch_hilight} shows an unrolled factor graph for a corpus containing two documents, two words per document and three topics. When we describe the algorithm below, we refer to one \textit{branch} of the LDA graph; this relates to one specific word, $n$, in a single specific document, $m$. The branch where $m =1$ and $n = 2$ is highlighted in Figure \ref{fig:LDA_branch_hilight}.
\subsection{Description of the ALBU LDA algorithm}
We first present our approach to the Bayesian parameter updating procedure for LDA, defining our notation for the hyperparameters, prior distributions and posterior distributions. Thereafter, using notation from Fig.~\ref{fig:LDA_branch_hilight}, we start at the word-topic Dirichlet distributions and present the factor update equations towards the topic-document Dirichlet distributions (forward sweep). After updating these Dirichlet distributions (by means of an approximation that is based on sampling) we present the update equations in the reverse direction (backward sweep). 
\subsubsection{General Algorithm for Bayesian parameter updating procedure for LDA}
The hyperparameters are typically chosen (as described in Section 4) to be values between 0 and 1 to favor sparse distributions; $\bm{\beta}_1$ up to $\bm{\beta}_{K}$ are the parameters of the prior Dirichlet word-topic distributions and $\bm{\alpha}_1$ up to $\bm{\alpha}_{M}$ correspondingly are the parameters of the prior Dirichlet topic-document distributions. 

The message passing procedure will update these prior Dirichlet distributions to the posterior distributions (during each epoch) which will simply result in an update to the Dirichlet parameters. We will use $\bm{\beta}'$ and $\bm{\alpha}'$ values respectively to denote these updated parameters.

During each epoch, the two internal distributions in each branch ($f_{Z_{m,n}}$ and $f_{W_{m,n}}$) operate on the latest posterior Dirichlet estimates -- with a small correction to prevent over-counting due to its own contribution to the posterior during the previous epoch. This is standard message passing protocol \cite{heskes2003stable, lauritzen1988local} and is implemented by subtracting its own increment to the posterior counts from the previous round of message passing. 

It is important to note that although we are presenting these equations in a factor graph form, our implementation uses a cluster graph architecture. Most readers will be familiar with the Sum-Product algorithm for factor graphs. We are using a slightly modified version of this (Loopy Belief Update) as described by Lauritzen-Spiegelhalter \cite{lauritzen1988local} where reverse direction messages are not excluded from the current message calculation, but rather cancelled.

When referring to parameters post adjustment (the corrected values), $\bm{\beta}^{''}$ and $\bm{\alpha}^{''}$ will be used. These internal distributions, in turn, update the posterior Dirichlet distributions. 

This process iterates until a convergence criterion is met. We can now define the initial factors in a specific branch:
\begin{itemize}
    \item Topic-Document Dirichlet factors, $f_{\theta_m}$, have an initial belief of \\ $p(\bm{\theta}_m; \alpha_{m,1},..., \alpha_{m, K}) = p(\bm{\theta}_m;\bm{\alpha}_m)$.
    \item Topic-Document Polya factors, $f_{Z_{m,n}}$, initially have a belief of $p(Z_{m,n};\bm{\theta}_m)$.
    \item Word-topic conditional Polya factors, $f_{W_{m,n}}$, initially have a factor belief of $p(W_{m,n}|Z_{m,n},\bm{\Phi})$.
    \item Word-topic Dirichlet set factors,  $f_{\Phi}$, have an initial factor belief of $p(\bm{\phi}_1, ...,\bm{\phi}_{K};\bm{\beta}_1, ...,\bm{\beta}_{K} ) = p(\bm{\Phi};\bm{B})$.
\end{itemize}

\subsubsection{Forward sweep of ALBU LDA}
During the forward sweep of each epoch, each word-topic Dirichlet factor node updates its respective word-topic Dirichlet variable node (initially, in the first epoch, this posterior Dirichlet is the same as the prior Dirichlet, parameterized by $\bm{\beta}^{'} = \bm{\beta}$). The Dirichlet variable node then sends a message to its respective conditional word-topic Polya factor node,  $f_{W_{m,n}}$. Initially this factor node contains a conditional distribution which is defined as a categorical distribution for each topic, $p(W_{m,n}|\bm{\phi}_{k}, Z_{m,n}=k) =\prod_v\phi_{k,v}^{[W_{m,n}=v]}$.

For each topic, we multiply this categorical distribution with the incoming message from the corresponding word-topic Dirichlet. The word-topic variable node (posterior Dirichlet) is defined as,
\begin{equation}
    V_{\phi_k}^{\text{post}}=\mathsf{Dir}(\bm{\phi}_{k};\bm{\beta}^{'}_{k})  =  \frac{\Gamma(\sum_{v}\beta^{'}_{k,v})}{\prod_v\Gamma(\beta^{'}_{k,v})} \prod_{v}\phi_{k,v}^{\beta^{'}_{k,v}-1},
    \label{eq:twvn}
\end{equation}

To update the conditional word-topic Polya distributions, we correct for the contribution of the message sent in the reverse direction by using adjusted $\bm{\beta}^{''}$ Dirichlet parameters (these adjusted values are calculated by summing and subtracting $\beta$ parameters as the multiplications and divisions involve Dirichlet and Polya distributions).

To get an update for the full posterior joint distribution at the factor node, $f_{W_{m,n}}$, we also need to multiply in the message from the variable node, $V_{Z_{m,n}}$. This message, $p(Z_{m,n};\bm{\alpha}^{''}_{m,n})$, is parameterized by adjusted $\bm{\alpha}^{''}_{m,n}$ topic-document Dirichlet parameters (due to message cancelling) for each word $n$ in each document $m$ and is derived in the backward sweep (Eq.~\ref{eq:pol_to_cpol}). The current posterior at $f_{W_{m,n}} = p(W_{m,n},Z_{m,n},\bm{\Phi};\bm{\alpha}'',\bm{B}'')$, and can be written as,

\begin{equation}
 f_{W_{m,n}}= \left\{
    \begin{array}{lr}
    p(Z_{m,n}=1)\frac{\Gamma(\sum_v\beta^{}_{1,v})}{\prod_{v}\Gamma(\beta^{}_{k,v})}\prod_v\phi_{1,v}^{[W_{m,n}=v]+\beta^{}_{1,v}-1} \\[1mm]
    \vdots &\\ [1mm]
    p(Z_{m,n}=K)\frac{\Gamma(\sum_v\beta^{}_{K,v})}{\prod_{v}\Gamma(\beta^{}_{k,v})}\prod_v\phi_{K,v}^{[W_{m,n}=v]+\beta^{}_{K,v}-1}.
\end{array}
\right.\label{eq:condpolyalda}
\end{equation}

 Note that the hyperparameters are unique for each branch (for specific ${m}$ and ${n}$) due to message refinement but we do not indicate this in the subscripts (for $\beta_{k,v}$ this would be $\beta_{v,Z_{\mathsf{m},\mathsf{n}}={k}}$) to avoid unnecessary notational complexity.
 
To update the topic-document Polya we use Equation \ref{eq:condpolyalda}. In LDA, the word is observed and this simplifies our update (for a single topic, $\mathsf{k}$) as follows,

\begin{align}
  \Tilde{\mu}_{f_{W_{m,n}|Z_{{m},{n}}=\mathsf{k}}\xrightarrow{}{Z_{m,n}}}
  &= \frac{p(Z_{m,n}=\mathsf{k})\Gamma(\sum_v\beta_{v,Z_{{m},{n}}=\mathsf{k}}) }{\prod_{v}\Gamma(\beta_{v,Z_{{m},{n}}=\mathsf{k}}-1)}\int_{\bm{\phi}_{\mathsf{k}}} \phi_{\mathsf{v},Z_{{m},{n}}=\mathsf{k}}^{[W_{{m},{n}}=\mathsf{v}]+\beta_{\mathsf{v},Z_{{m},{n}}=\mathsf{k}}-1}d \bm{\phi}_{\mathsf{k}}\notag\\
  &= \frac{p(Z_{m,n}=\mathsf{k})\Gamma(\sum_v\beta_{v,Z_{{m},{n}}=\mathsf{k}}) }{\prod_{v}\Gamma(\beta_{v,Z_{{m},{n}}=\mathsf{k}}-1)} \int_{\bm{\phi}_{\mathsf{k}}} \phi_{\mathsf{v},Z_{{m},{n}}=\mathsf{k}}^{\beta_{\mathsf{v},Z_{{m},{n}}=\mathsf{k}}}d \bm{\phi}_{\mathsf{k}}  \notag\\
  &= p_{\mathsf{k},\mathsf{v},m,n} \frac{\beta_{\mathsf{v},Z_{{m},{n}}=\mathsf{k}}}{\sum_v\beta_{v,Z_{{m},{n}}=\mathsf{k}}}. 
  \label{eq:marg_theta_cpolrev}
\end{align}

where $p_{\mathsf{k},\mathsf{v},{m},{n}}$ are the current topic proportions for each observed word and topic (per branch) such that $p_{\mathsf{k},\mathsf{v},{m},{n}} = p(Z_{m,n}=\mathsf{k})$. 

Equation \ref{eq:marg_theta_cpolrev}, however, is un-normalized (as indicated by the $\Tilde{.}$ ) and has a volume of $\text{vol}_{\mathsf{v},\mathsf{k}} = Z$. We need to normalize the messages for each topic by multiplying each $\Tilde{\mu}_{f_{W_{m,n}|Z_{{m},{n}}=\mathsf{k}}\xrightarrow{}{Z_{m,n}}}$ by a normalizing constant to give,

\begin{align}
 \mu_{f_{W_{m,n}|Z_{{m},{n}}=\mathsf{k}}\xrightarrow{}{Z_{m,n}}}
  &=  \frac{1}{Z}p_{\mathsf{k},\mathsf{v},{m},{n}} \frac{\beta^{''}_{\mathsf{v},Z_{{m},{n}}=\mathsf{k}}}{\sum_v\beta^{''}_{v,Z_{{m},{n}}=\mathsf{k}}}. 
  \label{eq:marg_theta_cpol}
\end{align}

We can now update each of the topic-document Polya factors (defined in Equation \ref{eq:fz}) by multiplying in the message contributions of each topic from its relevant conditional word-topic Polya distribution (after applying message cancelling). 

We can now update the topic-document Dirichlet distributions with our latest belief about $\bm{\theta}_{m}$ for each branch. To calculate the message from $f_{Z_{m,n}}$ to ${\theta_{m}}$, the contribution of $Z_{m,n}$ needs to be marginalized out. This results the following function,

\begin{align}
\mu_{f_{Z_{m,n}}\xrightarrow{}{\theta_{m}}}
 &=\sum_{Z_{m,n}} p(Z_{m,n},\bm{\theta}_{m};\bm{\alpha}^{}_{m,n})\notag\\
    &= \left(\frac{\Gamma(\sum_k\alpha^{}_{m,n,k})}{\prod_k\Gamma(\alpha^{}_{m,n,k})}\prod_k\theta_{m,k}^{\alpha^{}_{m,n,k}-1}\right)\left(\frac{\sum_k\alpha^{}_{m,n,k}\sum_kp_{m,n,k}\theta_{m,k}}{\sum_kp_{m,n,k}\alpha^{}_{m,n,k}}\right),
  \label{eq:dirich_true_msglda}
\end{align}

which is unfortunately not conjugate to a Dirichlet distribution. 

We know how to update the Dirichlet posterior if we observe values for $Z$ -- simply
increment the count of the corresponding $\bm{\alpha}^{}_{m}$ values. Because $Z$ is latent, however, the best we can do is to use the topic proportions to scale the respective $\bm{\alpha}^{}_{m}$ values before adding this scaled fractional count to the current $\bm{\alpha}^{}_{m}$. We validate this by sampling a large number of $Z$ values and averaging the counts. When doing this, the relative proportions are consistent with this fractional count approach and results in the update,
 
\begin{equation}
    \alpha^{'}_{m,k} = \sum_n\frac{p^{''}_{m,n,k}\alpha^{''}_{m,n,k}}{\sum_i p^{''}_{m,n,i}\alpha^{''}_{m,n,i}} + \alpha_{m,k},\label{eq:approxmessage}
\end{equation}
where $\bm{\alpha}^{''}_{m}$ are obtained by cancelling out the message in the reverse direction. It is interesting to note that this update equation is consistent with the message update that is obtained by using variational message passing (in the special case where the message does not use the expectation of the natural parameters but rather the true full distribution). This is not too surprising since the Sum-Product algorithm (as well as LBU) is variational in nature.

Equation \ref{eq:approxmessage} updates $\bm{\alpha}^{'}_{m}$ which can then be inserted the document topic Dirichlet to arrive at the posterior (for that epoch),
\begin{equation}
f^{\text{post}}_{\theta_m} = \mathsf{Dir}(\bm{\theta}_{m};\bm{\alpha}^{'}_{m}) =  \frac{\Gamma(\sum_k\alpha^{'}_{m,k})}{\prod_k\Gamma(\alpha^{'}_{m,k})} \prod_{k}\theta_{m,k}^{\alpha^{'}_{m,k}-1}.
\end{equation}

\subsubsection{Backward sweep}
In the first iteration, the posterior $\bm{\alpha}^{'}$ values are the same as the prior values. These are used to update the topic-document categorical distributions after adjusting for over-counting (because the reverse messages in the initial epoch are uninformative, we have $\bm{\alpha}^{''} = \bm{\alpha}^{'}$ in the first epoch) resulting in,
\begin{equation}
    f_{Z_{m,n}}
    =\frac{\Gamma(\sum_k\alpha^{''}_{m,n,k})}{\prod_{k}\Gamma(\alpha^{''}_{m,n,k})} \prod_{k}\theta_{m,k}^{\mathbb{[}Z_{m,n}=k\mathbb{]}+\alpha^{''}_{m,n,k}-1}
    \label{eq:fz}.
\end{equation}
To calculate the message to be sent to $f_{W_{m,n}}$ from the left, we need to integrate out $\bm{\theta}_{m}$, giving, 
\begin{equation}
p(Z_{m,n}=k;\bm{\alpha}^{''}_m) =  \frac{\alpha^{''}_{m,n,k}}{\sum_k\alpha^{''}_{m,n,k}},
\label{eq:pol_to_cpol}
\end{equation}
which is unique for each branch due to the message cancelling adjustment to the hyperparameters. We can now update Equation \ref{eq:condpolyalda} with the latest version of $p(Z_{m,n}=k;\bm{\alpha}^{''}_m)$ from Equation \ref{eq:pol_to_cpol}.

To complete our reverse sweep we observe $W_{m,n} = \mathsf{v}$ and calculate the the latest topic proportions from Equation \ref{eq:condpolyalda} (for each topic, $\mathsf{k}$),

\begin{align}
  \Tilde{\mu}_{f_{W_{m,n}|Z_{{m},{n}}=\mathsf{k}}\xrightarrow{}{Z_{m,n}}}
  &= p_{\mathsf{k},\mathsf{v},{m},{n}} \frac{\beta_{\mathsf{v},Z_{{m},{n}}=\mathsf{k}}}{\sum_v\beta_{v,Z_{{m},{n}}=\mathsf{k}}}. 
  \label{eq:marg_theta_cpolrev2}
\end{align}
We can now update the topic proportions for each word within each word-topic Dirichlet (after adjusting for over-counting). At the word-topic Dirichlet variable nodes, these proportions are simply added as partial counts to the corresponding Dirichlet distribution's prior values to give the latest updated posterior Dirichlet parameters from all branches within the document, 
\begin{equation}
    \beta^{'}_{\mathsf{k},\mathsf{v}} = \sum_m\sum_n\frac{p'_{\mathsf{k},\mathsf{v},{m},{n}}}{\sum_k p_{\mathsf{k},\mathsf{v},{m},{n}}} + \beta_{\mathsf{k},\mathsf{v}},
\end{equation}

which are used to update the respective posterior word-topic 
Dirichlet Distributions given in Eq.~\ref{eq:twvn}. It is important to note here that each $p_{\mathsf{k},\mathsf{v},{m},{n}}$ contains both updated and adjusted $\alpha$ and $\beta$ hyperparameters. Note also that unlike when the topic-document Dirichlet distributions are updated, there is no approximation in this Dirichlet update. This message differs from the VMP message in that the VMP message here is independent of the $\beta$ hyperparameters.

In this section we have given the update equations for the ALBU algorithm. From an implementation point of view, by inspecting the equations, it is clear that all updates are simply subtraction, multiplication and division of vectors. Also, unlike VMP and VB, no complex functions such as the Digamma functions are used, allowing for an improved execution time. Zeng also notes this advantage in his collapsed Sum-Sum LDA algorithm \cite{zeng2012learning}.

We will now present the evaluation process that we employed to test our implemented algorithm.

\section{Our testing methodology}
We compare the performance of three algorithms, namely ALBU, collapsed Gibbs sampling and VB. Four text corpora are used for this comparison and are all considered small  (they contain few terms per document and also contain a relatively small number of documents). 

Because the ground truth distributions are not known, we also include two simulated data sets where 20 corpora are generated for each data set. We then use the ground truth distributions from which the corpora are generated to more accurately evaluate these algorithms.
\subsection{Metric selection}
In supervised learning models, the ability of a trained model to predict a target variable is evaluated using a test set. Evaluating the performance of unsupervised learning algorithms such as LDA \cite{mcauliffe2008supervised}, is less straightforward and a measure of success needs to be defined. Held-out perplexity \cite{chang2009reading} has been the most popular evaluation metric for topic models.  However, a large-scale human topic labeling study by Chang et al. \cite{chang2009reading}
demonstrated that low held-out perplexity is often poorly correlated with interpretable latent spaces. Coherence on the other hand, has been shown to be highly correlated with human interpretability of topics \cite{roder2015exploring}. 

In a comprehensive study of multiple coherence measures \cite{roder2015exploring}, the $C_{\mathsf{V}}$ coherence score had the highest correlation with human topic ratings. This measure is a combination of three measures: the indirect cosine measure, the Boolean sliding window and the normalized point-wise mutual information score, $C_{\mathsf{V}}$,  which performed almost as well as the $C_{\mathsf{V}}$ score.  Other well-known coherence measures evaluated in their analysis include, in order of performance are  $C_{\mathsf{UCI}}$ and $C_{\mathsf{UMass}}$ \cite{roder2015exploring}. We therefore use the $C_{\mathsf{V}}$ score and the simpler
$C_{\mathsf{NPMI}}$ score to evaluate the topic modeling of the text corpora. Using these measures in conjunction with one another allows for selection of topics where both scores indicate good performance. There are cases where only one metric performs well and the other does not, this usually indicates performance that is not as good as the higher scoring metric leads one to believe.

The evaluation metrics mentioned above are not without drawbacks. Coherence measures take only the top words per topic into account, and not the full distributions over topics. Consequently, much detail of the learnt distributions is discarded. Perplexity may also not give sufficiently fine-grained resolution: Minka and Lafferty \cite{minka2002expectation} address similar concerns. They demonstrate that held-out word perplexity for two different models can be almost identical but when inspected  (using simulated data where the word-topic and topic-document distributions are known), large performance differences are seen \cite{minka2002expectation}. 

To evaluate our algorithm based on a more fine-grained metric, we implement a corpus simulation system in which the topic-document and word-topic distributions are known. A distance measure (forward Kullback-Liebler divergence (KLD)) is then used to compare the learnt distributions with the true distributions.

The word-topic and topic-document distributions are Dirichlet distributions. One can easily calculate the forward KLD between two Dirichlet distributions. 

Unfortunately, the Gensim implementation (to evaluate the VB algorithm) allows access only to the mean of these Dirichlet distributions, not to the distributions
themselves. In LDA the mean of these distributions is, in fact, the probability of finding a word in a topic. We therefore calculate the forward KLD between the actual mean word-topic distributions, $p$, and approximate mean word-topic distribution, $q$ for each topic:
\begin{equation}
    \bm{KL}(p\parallel q) = \sum_i p_i \ln{\frac{p_i}{q_i}}
\end{equation}

The average KLD over all topics is taken to be the error for each model.

\subsection{Text Corpora}
To evaluate the algorithms on text corpora, three text sources are chosen, namely (1) a selection of 20000 Covid-related tweets, (2) \textit{Covid Tweets}, the Bible verses from key parts of the new testament, \textit{Bible Verses}, and (3) the widely used 20 Newsgroups corpus, \textit{20 Newsgroups}. Each can be considered a small corpus, in terms of both the number of documents they contain and the text lengths of each document. 
 
 \begin{figure}
  
    \begin{subfigure}{0.49\textwidth}
  \centering
  \includegraphics[width=\linewidth]{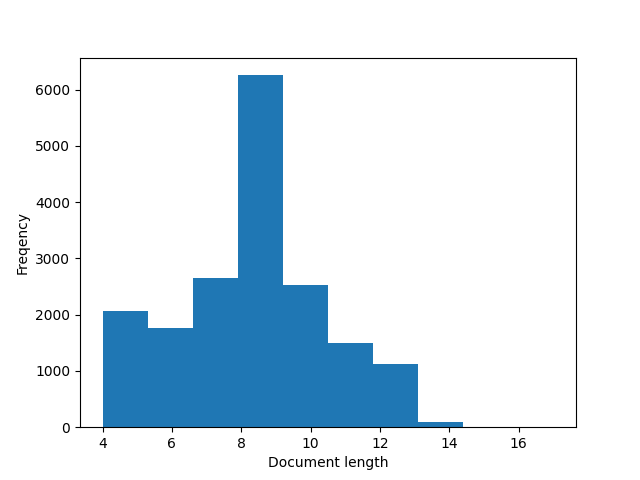}
  \caption{Document length for the Covid Tweets. Because the original texts (tweets) are so short, very few words remain in each document after preprocessing. Documents with fewer than 4
words are excluded from the analysis. \label{doclengthcovid}}
  \end{subfigure}
\hfill
  \begin{subfigure}{0.49\textwidth}
  \centering
  \includegraphics[width=\linewidth]{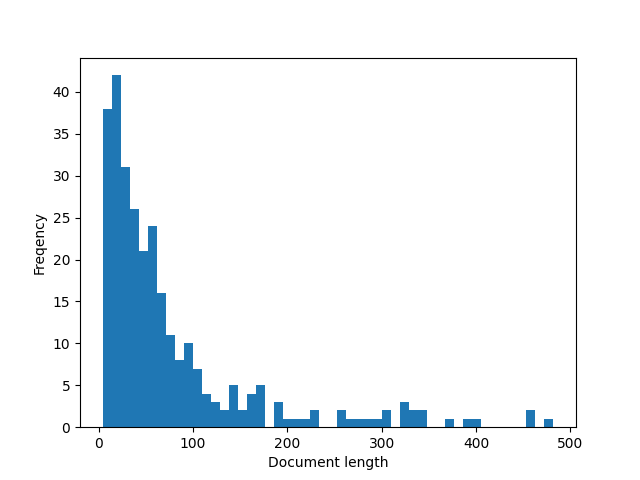}
  \caption{Document length for the 6 Newsgroups corpus. We cut off the plot at 500 documents because only very few documents had lengths between 500 and 3400. The distribution for the 20 Newsgroups corpus is similar. \label{fig:ARTfreq20newssmall}}
  \end{subfigure}
  \caption{Frequency of document lengths for the Covid and 6 Newsgroup corpora. Statistics of the document lengths for the corpora can be seen in Table \ref{table:priors}.\label{fig:freq}}
\end{figure}
\begin{figure}
  
    \begin{subfigure}{0.48\textwidth}
  \centering
  \includegraphics[width=\linewidth]{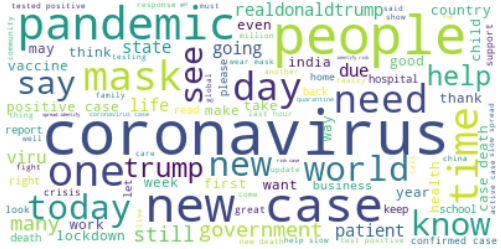}
  \caption{Wordcloud for Covid Tweets corpus. The token \textit{covid} has been removed since it is present in every tweet. The \textit{coronavirus} token has been retained in the corpus.
  .\label{fig:wca2}}
  \end{subfigure}
\hfill
  \begin{subfigure}{0.48\textwidth}
  \centering
  \includegraphics[width=\linewidth]{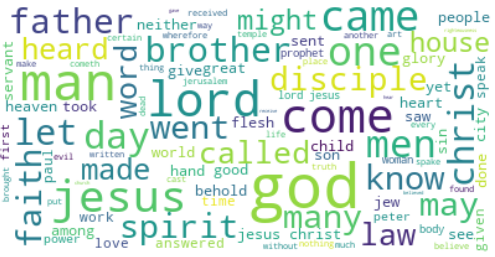}
  \caption{Wordcloud for Bible verses from the Gosples, Acts and the Epistles - books that are known to have overlapping themes. \label{fig:wcmored}}
  \end{subfigure}
  \caption{Word clouds are based on the frequency of individual words occurring in the corpus. In (a) we show the Covid Tweets corpus and (b) the Bible verses corpus.\label{fig:wccov}}
\end{figure}

In all of the text corpora we remove all characters that are not in the English alphabet. Standard text processing techniques are then applied such as lemmatisation and stopword removal. We also remove some specific additional stopwords for each corpus, such as \textit{corona} from \textit{Covid Tweets} because 
this is the word used to identify the selection of tweets. Documents shorter than 4 tokens in length after initial preprocessing are discarded.

Because we do not know the true number of topics for any of these corpora, we create a fourth corpus which is simply a small subset of the 20 Newsgroups corpus, \textit{6 Newsgroups}, where we select only $6$ of the $20$ newsgroups and aim to identify these newsgroups as topics. The newsgroup names of these $6$ newsgroups are \textit{recmotorcycles}, \textit{comp.graphics}, \textit{sci.med}, \textit{sci.space}, \textit{talk.politics.mideast} and \textit{talk.religion.misc}. To make the problem harder for the algorithms to solve, we also limit the number of documents per topic to 50, leaving us with a very small corpus containing only 300 documents. A summary of the corpus statistics can be seen in Table ~\ref{table:priors}.

 \begin{table}
 \small
  \caption{Table showing text corpus statistics and $\alpha$ and $\beta$ values chosen for evaluation.}
  \label{table:priors}
  \begin{tabular}{ccccc}
      \toprule

      & Covid Tweets       & Bible Verses  & 20 Newsgroups & 6 Newsgroups  \\
          \midrule
    Number of documents          & $18951$ & $7554$  & $300$  & $6996$ \\
    Vocabulary size          & $7785$ & $2627$  & $18035$ & $5631$  \\
    Total topics           & $8-13$ & $9-16$  & $11-21$  & $6$   \\
    Document length range          & $4-16$  & $4-26$ & $4-4537$ & $4-4537$  \\
    Mean document length          & $8$  & $8$ & $21$  & $17$ \\
    Coherence window         & $15$ & $15$ & $50$  & $15$  \\
    $\alpha$          & $0.1$ & $0.1$ & $0.1$ & $0.5$   \\
    $\beta$          & $0.1$  & $0.1$ & $0.1$ & $0.5$  \\
      \bottomrule
  \end{tabular}

\end{table}

\begin{figure}
  
    \begin{subfigure}{0.48\textwidth}
  \centering
  \includegraphics[width=\linewidth]{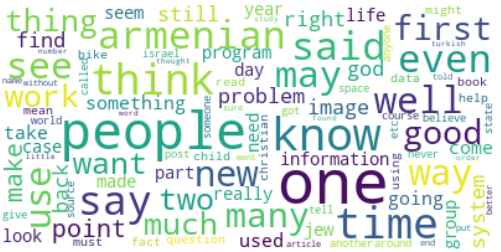}
  \caption{Wordcloud for 20 Newsgroups corpus. We see that \textit{people} and \textit{one} are two of the most common words. There are many general words not obviously relating to a specific topic, except possibly \textit{armenian}, which belongs primarily to the \textit{talk.politics.mideast}  group.\label{fig:wca}}
  \end{subfigure}
\hfill
  \begin{subfigure}{0.48\textwidth}
  \centering
  \includegraphics[width=\linewidth]{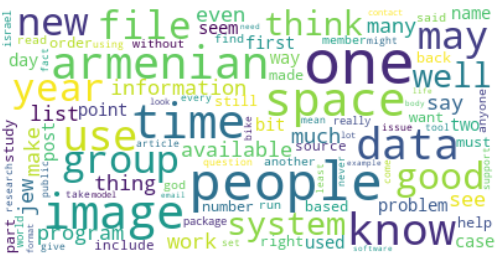}
  \caption{Wordcloud for 6 Newsgroups corpus. Here some of the 6 newsgroups are clearly represented by the most frequent words such as \textit{space} from \textit{sci.space}, \textit{image} and \textit{file} from \textit{comp.graphics} and \textit{armenian} from \textit{talk.politics.mideast}. \label{fig:wcs}}
  \end{subfigure}
  \caption{Word clouds are based on the frequency of individual words occurring in a corpus. In (a) we show the full 20 Newsgroups corpus and in (b) the smaller corpus using only 6 of the newsgroups.\label{fig:wc20}}
\end{figure}

\subsection{Simulated corpora}
The corpus is generated as follows: (a) generate word-topic distributions, (b) generate topic-document distribution, (c) for each document, repeatedly choose a topic from its topic-document distribution, and then a word from this word-topic distribution until the document is populated with the required number of words.

\begin{figure}
  
    \begin{subfigure}{0.32\textwidth}
  \centering
  \includegraphics[width=\linewidth]{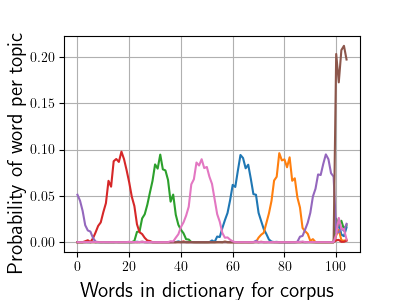}
  \caption{Extracted word-topic distributions for a corpus containing 200 documents ($M = 200$) in the smaller simulated data set. The 7th topic is the stop words topic and contains words that are found in all documents. }
  \end{subfigure}
    \hfill
  \begin{subfigure}{0.32\textwidth}
  \centering
  \includegraphics[width=\linewidth]{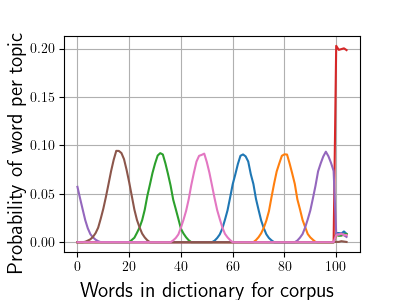}
  \caption{True word-topic distributions for a corpus containing 5000 documents ($M = 5000$) from the same data set as (a). It is evident that because there are more samples, the underlying word-topic distributions are learnt more exactly.}
  \end{subfigure}
  \hfill
  \begin{subfigure}{0.32\textwidth}
  \centering
  \includegraphics[width=\linewidth]{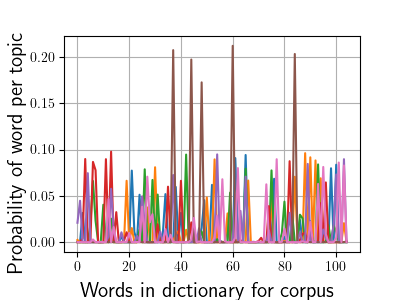}
  \caption{Shuffled true word-topic distributions for the corpus shown in (a). This simply illustrates that no reliance can be placed on the sequence of words within a topic or the fact that adjacent topics are significantly more likely to share words.}
  \end{subfigure}
  \caption{In (a) we present the word-topic distributions extracted from a corpus from the smaller simulated data set where $M = 200$, in (b), a corpus from the same data set where $M = 5000$ and in (c) a shuffled version of (a) to show that the sequential nature of words in topics is simply so that one can easily evaluate performance by visually inspecting the approximate versus actual distributions.\label{fig:mini1}}
\end{figure}

To the topics mentioned above, we add an additional topic, non-overlapping with the others, but occurring in all documents. The addition of these function (or stop) words makes the task of learning the word-topic and topic-document distributions significantly more challenging. This is one of the challenges when applying LDA to text corpora and it is typically handled by applying preprocessing techniques before running LDA (such as removing stop words or using the TF-IDF) \cite{wallach2009rethinking} or by post-processing (removing context-free words after extracting topics \cite{minka2002expectation}). By including these function words, we aim to make our simulations more difficult and realistic. 

An added benefit of using simulated data is that we can visualise the performance by using words to be numbers and assigning the word-topic probabilities in such a way that when ordered sequentially, the topics can be visually inspected. This can be seen in Fig.~\ref{fig:mini1}.

It is important to note that each time we chose documents based on corpus settings (such as number of documents per topic, words per document, vocabulary length, etc.) one can sample as many corpora as one likes from a single setting. The more are present documents per corpus, the greater the similarity between generated corpora, and the easier for LDA to learn the underlying distributions.

We utilize two simulated data sets (using 2 unique sets of corpus settings as shown in table \ref{table:simulated}) each consisting of 20 groups of corpora. These corpora are small by real-word text topic extraction standards, but this does not make them easier for LDA to learn; smaller data sets contain less information and consequently inference regarding the underlying generating distribution is more difficult.

The parameters of the smaller data set was inspired by a real data set in the education domain. The larger data set is created with the aim of being a more difficult problem to solve given the higher number of topics per document ($6$ topics), with documents nevertheless being relatively short ($120$ words).

 \begin{table}
  \caption{Table showing simulated corpus statistics as well as chosen $\alpha$ and $\beta$ values for algorithms evaluation.}
  \label{table:simulated}
  \begin{tabular}{ccccc}
  
    \toprule

           & Smaller Simulated & Bigger Simulated \\
                \midrule
  
    Number of documents          & $50, 100, 200, 300,500,5000$  & $100,200,300,500$ \\
    Vocabulary size          & $100$  & $500$  \\
    Topics per document          & $3$ & $6$ \\
    Total topics           & $7$ & $10$   \\
    Document length          & $100$  & $120$  \\
    Coherence window         & $15$  & $10$ \\
    $\alpha$          & $0.5$  & $0.1$  \\
    $\beta$          & $0.5$  & $0.1$  \\
   \bottomrule
  \end{tabular}
\end{table}

\subsection{Hyperparameter selection}

\begin{figure}[ht]
  
    \begin{subfigure}{0.49\textwidth}
  \centering
  \includegraphics[width=\linewidth]{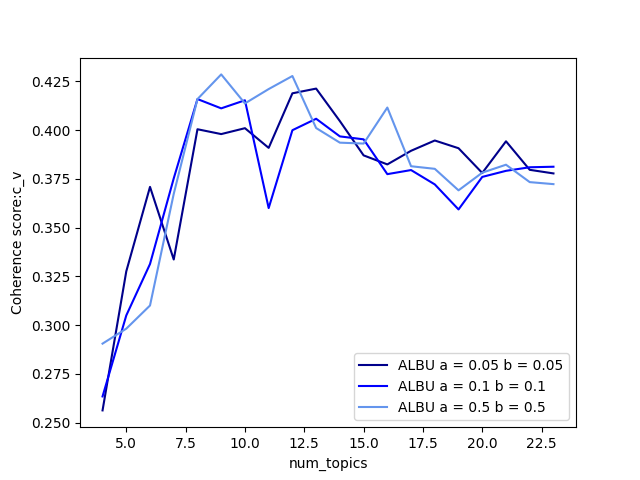}
  \caption{$C_{\mathsf{V}}$ scores over different hyperparameters for the Covid Tweets corpus.\label{fig:ARTalbuHyperCV}}
  \end{subfigure}
\hfill
  \begin{subfigure}{0.49\textwidth}
  \centering
  \includegraphics[width=\linewidth]{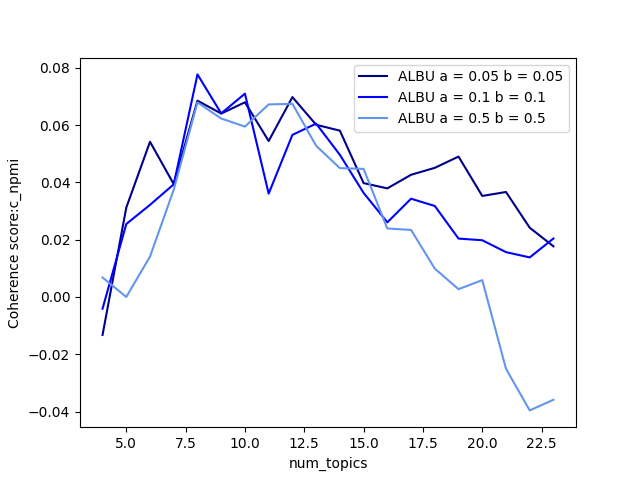}
  \caption{$C_{\mathsf{NPMI}}$ scores over different hyperparameters for the Covid Tweets corpus. \label{fig:ARTalbuHyperNPMI}}
  \end{subfigure}
  \caption{Hyperparameter selection based on Coherence for the Covid corpus. Both scores are taken into account to select the hyperparameters to use for algorithm evaluation for each corpus. In this case $\alpha = \beta = 0.1$ is the best setting. \label{fig:ARTalbuHyper}}
\end{figure}

 Standard practice has been to choose the same hyperparameter settings over all data sets and algorithms, popular choices being $\alpha = \beta = 0.1$ \cite{mukherjee2009relative} and $\alpha = \beta = 0.01$  \cite{zeng2012learning}. It has recently been shown that this is not optimal since the choice of the hyperparameters for the topic-document and word-topic Dirichlet priors has an impact on LDA performance \cite{asuncion2009smoothing}. An alternative is to learn the hyperparameter values based on a subset of the data, as described by Minka \cite{minka2002expectation, minka2000estimating}. Asunction et al. \cite{asuncion2009smoothing}, shows that using a grid search over a range of hyperparameters for each algorithm over the entire data set allows for an even better choice in hyperparameters, should the computational cost be worth it. 
 
 We choose a list of standard hyperparameter settings and evaluate their performance based on coherence metrics. An example of this is shown in Figure \ref{fig:ARTalbuHyper}. The final hyperparameters choices for each data set can be seen in Tables \ref{table:priors} and \ref{table:simulated}.

\subsection{Epochs and iterations}
\begin{figure}[ht]
  
    \begin{subfigure}{0.49\textwidth}
  \centering
  \includegraphics[width=\linewidth]{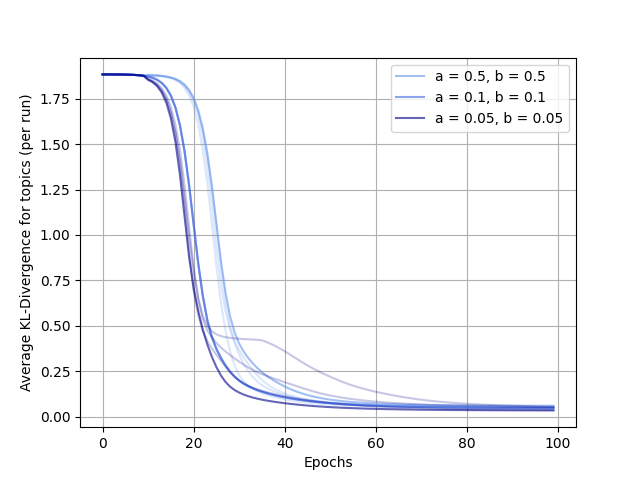}
  \caption{KLD convergence over different hyperparameters for a selection of corpora in the small simulated data set.\label{fig:epARTalbuHyperCV}}
  \end{subfigure}
\hfill
  \begin{subfigure}{0.49\textwidth}
  \centering
  \includegraphics[width=\linewidth]{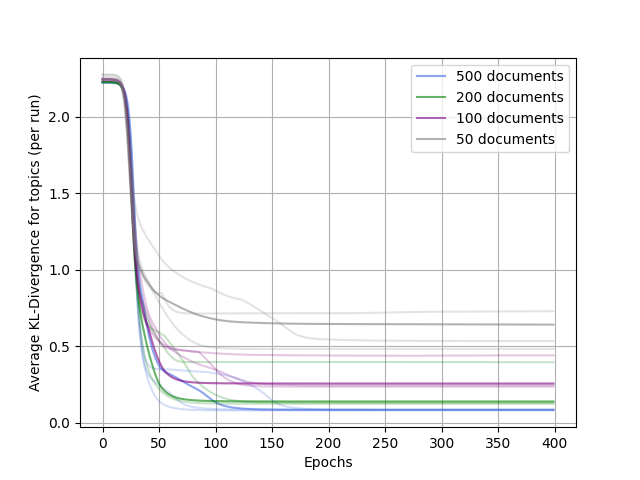}
  \caption{KLD convergence over different numbers of documents for a selection of corpora in the bigger simulated data set. \label{fig:epARTalbuHyperNPMI}}
  \end{subfigure}
 \caption{Epoch selection of simulated data based on KLD for ALBU. Algorithms converge at different rates based on hyperparameter settings as well as corpus settings such as number of documents, number of topics per document etc. \label{fig:ARTalbuHyperep}}
\end{figure}

Based on initial convergence tests, we fix the number of epochs for both ALBU and VB to 150 epochs for the text data sets. For the simulated data sets (where our results show the deviation from ground truth), we reduce the epochs to 70 for the smaller data set and increase the epochs to 200 for the 100 documents corpus for ALBU (based on results shown in Figure \ref{fig:ARTalbuHyperep}). For VB we use 150 epochs for all runs. Performance is significantly worse at 70 epochs for VB even on the smaller data set but shows no improvement at 200 epochs for either data set. 

For collapsed Gibbs sampling, 5000 samples are used after an initial 2000 iterations with rather poor results. This is significantly more than the 2000 samples than recommended in the Python package PyPi \cite{pypi} and 1000 as used by Zeng et al \cite{zeng2012learning}.

We test VB and ALBU only once per topic number on the text data sets, because the results for consecutive iterations are virtually identical. In the case of the collapsed Gibbs sampling
algorithm, however, the results vary (sometimes considerably) between iterations. We therefore follow the practice of Zeng et al. \cite{zeng2012learning} by performing multiple iterations per number of topics, and computing the average coherence value over these iterations.

For each simulated data set, we generate 20 corpora for each $M$ (number of documents). We refer to each repetition as a \textit{run}. In each run, topics are extracted by each algorithm, which are then compared to the true topics.

For the text data sets the $C_{\mathsf{V}}$ (a) and $C_{\mathsf{NPMI}}$ coherence scores are calculated for each topic within a topic range of $K = 4$ to $K = 16$ for the $\textit{Covid Tweets}$ and $\textit{Bible Verse}$ corpora, $K = 2$ to $K = 25$ for the $\textit{20 Newsgroups}$ corpus, and $K = 2$ to $K = 10$ for the $\textit{6 Newsgroups}$ corpus. 

\section{Results}
\label{sec:results}

In this section we first analyse the coherence results for the text corpora and then present the results for the simulated corpora. We utilise both coherence scores to select the best topic count range and then select a specific $K = k$ (number of topics) to inspect. Coherence measures have been shown to correlate with human topic interpretability. However, the best way of evaluating the intelligibility of and consistency within
individual topics is by visual inspection. For the \textit{Covid Tweets} and \textit{Bible Verses} corpora we aim to extract as few topics as possible out of the usable range. For the \textit{6 Newsgroups} corpus, we choose 6 topics since we want to determine how well the original newsgroup themes can be extracted. For the \textit{20 Newsgroups} data set we choose 13 topics (we see that many of the 20 newsgroups have common themes can are extracted as coherent topics). 

\subsection{Covid Tweets and Bible Verses}

Tweets are short messages that are posted by users on the Twitter platform and are typically only a sentence or two in length. The bible is divided into book types, books, chapters and verses. Verses are very short in length (similar to tweets). There is need in theology to find verses that cover the same topic. We select a group of similar biblical book and use each verse as a document in the corpus. We used this corpus because of its very short documents and limited vocabulary which make it difficult for algorithms to extract coherent topics. We expect coherence values to be low for such short documents (tweets and bible verses), especially with such small corpora because there is less information available from which to learn interpretable topics. 

\begin{figure}[ht]
  
    \begin{subfigure}{0.49\textwidth}
  \includegraphics[width=\linewidth]{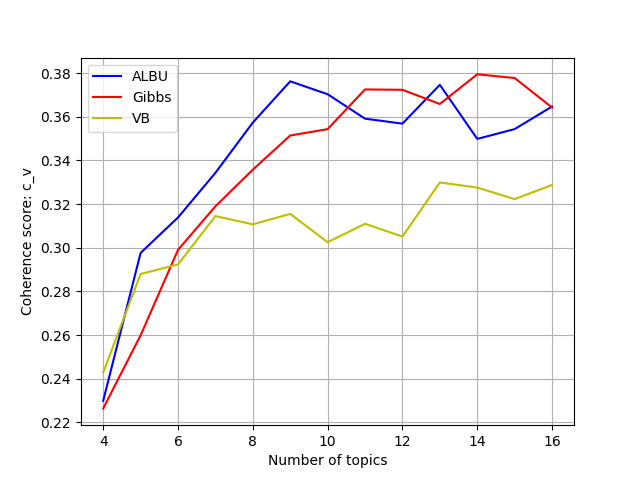}
  \caption{All three topics prefer topics from $9$ to $16$ in this metric with ALBU and Gibbs outperforming VB over the full range.\label{ARTbibleCV}}
  \end{subfigure}
\hfill
  \begin{subfigure}{0.49\textwidth}
  \includegraphics[width=\linewidth]{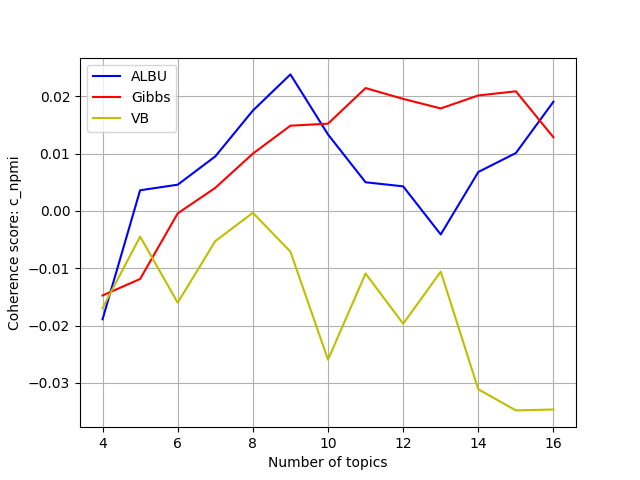}
  \caption{ALBU clearly prefers $9$ and $16$ topics while Gibbs is stable over range from $11$ to $15$. \label{fig:ARTbibleNPMI}}
  \end{subfigure}
  \caption{Coherence scores for the Bible verses corpus. We show the $C_{\mathsf{V}}$ scores in (a) and $C_{\mathsf{NPMI}}$ scores in (b). In both cases ALBU extracts topics best at $9$ and $16$ topics, while Gibbs prefers $10$ to $15$ topics. VB struggles to extract coherent topics.\label{fig:ARTbible}}
\end{figure}

\begin{figure}[ht]
    \begin{subfigure}{0.49\textwidth}
  \centering
  \includegraphics[width=\linewidth]{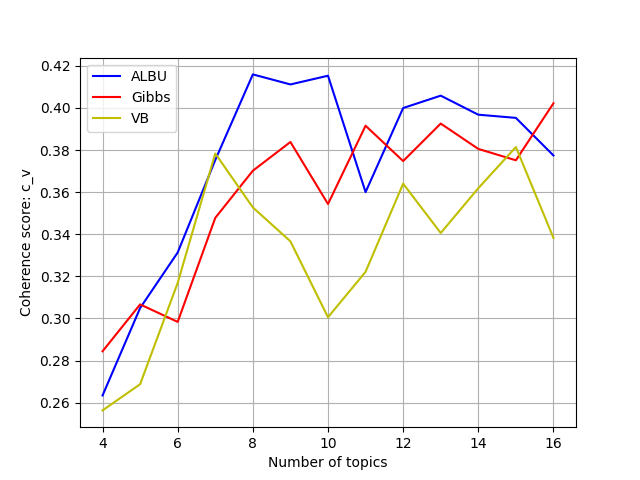}
  \caption{ALBU outperforms the others at all topic numbers except for Gibbs at $11$ and $16$ topics.\label{ARTcovidCV}}
  \end{subfigure}
\hfill
  \begin{subfigure}{0.49\textwidth}
  \centering
  \includegraphics[width=\linewidth]{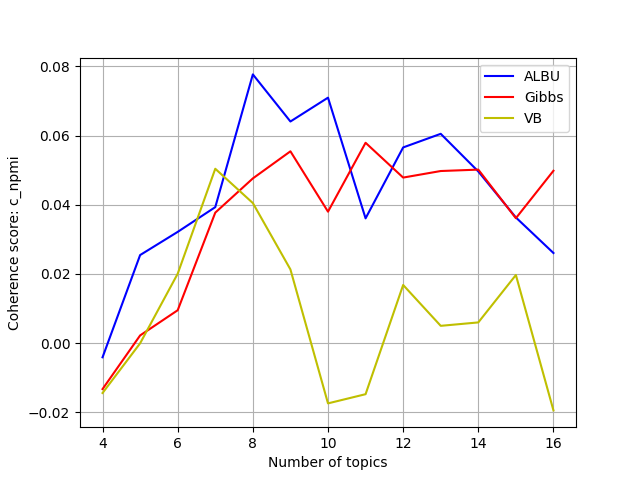}
  \caption{VB only extracts somewhat coherent topics at $K = 7$. \label{fig:ARTcovidNPMI}}
  \end{subfigure}
  \caption{Coherence scores for the Covid corpus.  We show the $C_{\mathsf{V}}$ score in (a) and $C_{\mathsf{NMPI}}$ in (b). We can see that ALBU extracts topics well at $7$ and $16$ topics while Gibbs prefers $10$ to $15$ topics. VB struggles to extract coherent topics except for at $K = 7$.\label{fig:ARTcovid}}
\end{figure}

From Figure \ref{fig:ARTbible} and Figure \ref{fig:ARTcovid} we can see that VB performs more poorly than do the ALBU and Gibbs sampling algorithms -- this is especially evident where the NPMI Metric is used to evaluate topic coherence. Visual examination of both of Figure \ref{fig:ARTcovid}'s plots indicates that the greatest
number of topics extracted by ALBU occurs where there are from $9$ to
$16$ topics, while the Gibbs sampling algorithm performs best at $10$ to
$1$ topics. In the case of VB, coherence scores are consistently lower
for most topic numbers. Indeed, in this analysis VB struggles to
extract coherent topics. 

Because all of the verses in the corpus are focused on similar topics, we aim to extract as few topics as possible, while maintaining acceptable topic coherence. The topics extracted by ALBU at  $K = 9$ are the most coherent; confirmed by coherence scores Figure \ref{fig:ARTbible}) as well as by inspection of the top $10$ words extracted per topic. In collapsed Gibbs sampling, when visually inspecting the topics, we see some duplication; for example, two topics contain the words \textit{God}, \textit{Christ} and \textit{Lord} and two additional topics contain the words\textit{God} and \textit{father}. ALBU extracts these as only two unique topics -- a more useful result.

For the Covid corpus we choose 
$K = 8$ for ALBU and $K = 9$ for the collapsed Gibbs sampling algorithm. By inspecting $6$ of the $8$ Covid topics extracted by ALBU in Table \ref{table:albuCOVID}, we see coherent, relevant topics such as mask wearing and testing. The extracted topics by collapsed Gibbs sampling are very similar to these especially with $K = 9$. VB extracts significantly fewer coherent topics; for example, two of the topics relate to both Donald Trump and children (\textit{Trump}, \textit{school}, and \textit{America} were top words in the one, and \textit{realdonaldtrump}, \textit{American} and \textit{school} in another topic). We see a similar trend for the Bible Verses topics over the three algorithms. 

\begin{table}[ht]
\tiny
  \caption{Top 15 words in 6 of the topics extracted from the Covid tweets corpus where $K = 8$ for ALBU.}
  \label{table:albuCOVID}
  \centering
  \resizebox{0.9\textwidth}{!}{%
  \begin{tabular}{ccccccc}
    \toprule

masks  & children  & cases   & trump & spread  & testing \\
\midrule
mask  & like  & case             & trump          & help  & test\\
people  & get      & new        & realdonaldtrump  & health  & positive\\
face  & school         & death     & via           & spread  & hospital\\
make  & back       & total      & american          & risk  & patient\\
wear  & know      & day         & every             & even  & testing\\
home  & want         & india     & give           & symptom  & tested\\
safe  & would  & number         & sign           & daily  & need\\
life  & going    & report       & support       & earth  & say\\
keep  & see        & today     & month            & self  & minister\\
stay  & still     & last         & change        &  slow  & chief\\
social & think  &  positive     & worker         &  community  & pradesh\\
one & thing     &  reported     & government     &  identify  & doctor\\
wearing &kid    &  update       & need      &  sooner  & govt\\

everyone & one  & hour          & million     &  long  & care\\
please & look   & confirmed     & america   &  impact  & police\\
    \bottomrule
  \end{tabular}}
\end{table}

\subsection{Newsgroup corpora}

\begin{figure}[ht]
 
    \begin{subfigure}{0.49\textwidth}
  \centering
  \includegraphics[width=\linewidth]{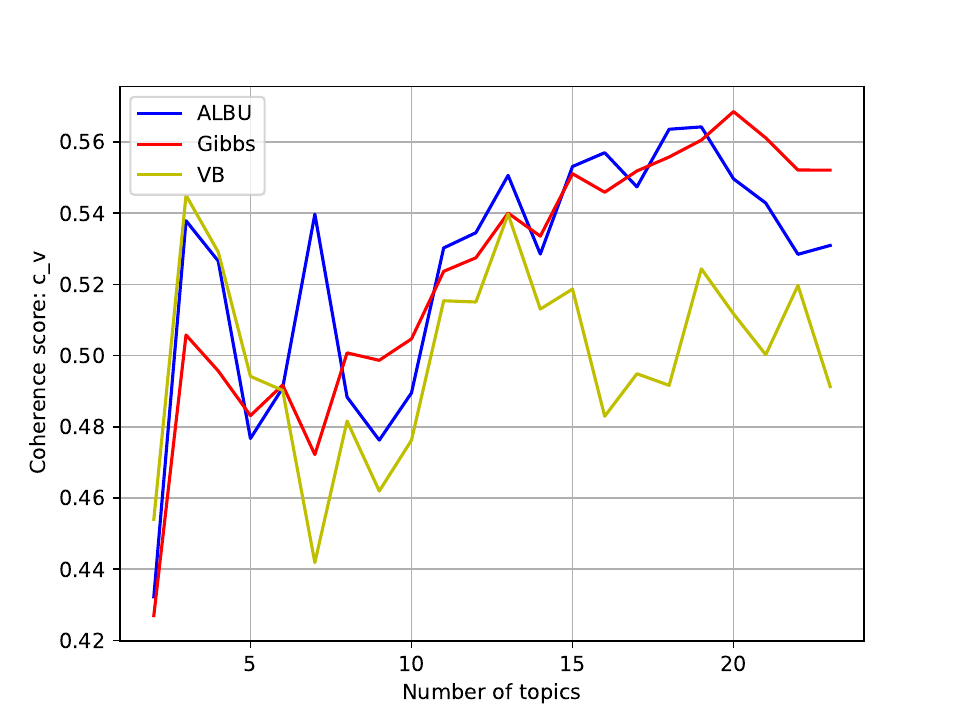}
  \caption{$C_{\mathsf{v}}$ scores for the three algorithms are similar for $K = 13$ which indicates that this is a good number of topics to inspect. The best score of topics over all algorithms over all $K$ is Gibbs at $K = 20$.\label{fig:ART20newsCVa}}
  \end{subfigure}
\hfill
  \begin{subfigure}{0.49\textwidth}
  \centering
  \includegraphics[width=\linewidth]{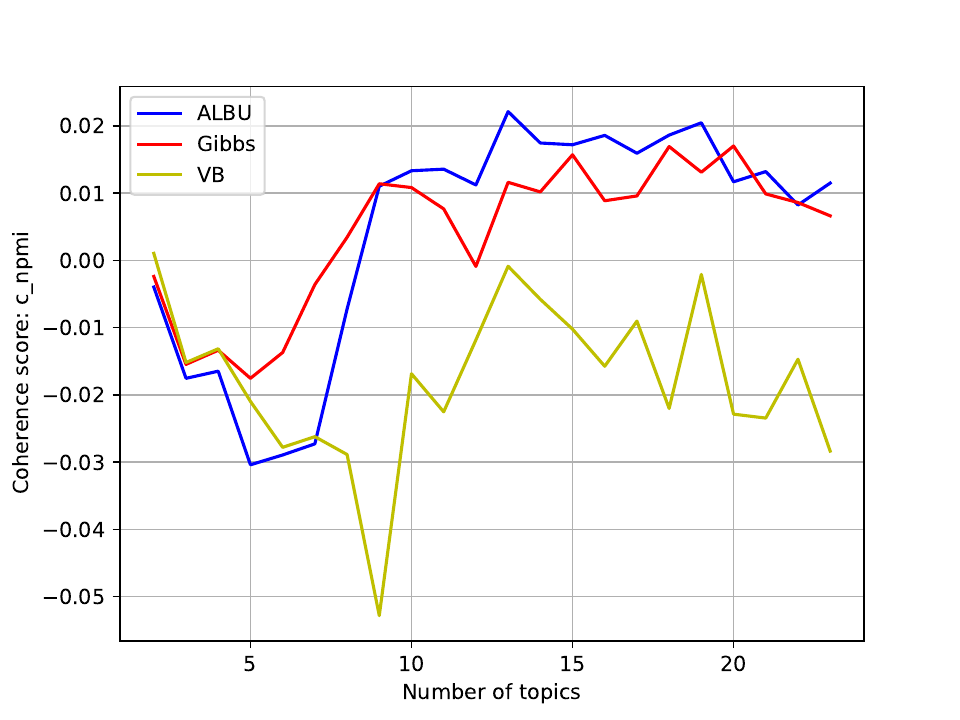}
  \caption{$C_{\mathsf{NPMI}}$ scores indicate that ALBU extracts topics better than the other algorithms with an exception of $K = 20$ where Gibbs performs best. The best score of topics over all algorithms over all $K$ is ALBU at $K = 13$. \label{fig:ART20newsNPMIb}}
  \end{subfigure} 
  \caption{Coherence scores for 20 Newsgroups. For both coherence metrics ALBU and Gibbs perform similarly with ALBU being marginally better at extracting topics. While still achieving reasonably good $C_{v}$ scores VB does significantly worse than the other two for $K > 14$.}
\end{figure}

We choose $K = 13$ since all three algorithms perform well for this number of topics with $C_{\mathsf{V}}$ scores of $0.54$ for both Gibbs and VB and of $0.55$ for ALBU (Figure \ref{fig:ART20newsCVa}). The $C_{\mathsf{NPMI}}$ score shows that Gibbs performs better than does VB and that ALBU outperforms both (Figure \ref{fig:ART20newsNPMIb}). Inspection of the actual words within the topics (Table \ref{table:20news}) extracted by ALBU, indicates that this is a sensible result. Many of the topics can be directly mapped to a specific newsgroup, as can be seen in Table \ref{table:20news} where $6$ of the $13$ topics are shown.

However, some topics contain words that seem to come from multiple newsgroups. This is to be expected, since some of the newsgroups address similar subjects. For example, the topic to which we have assigned the label \textit{hardware} contains words that are predominantly occur in two of the newsgroups, namely 
\textit{comp.sys.ibm.pc.hardware} and  \textit{comp.sys.mac.hardware}. This is the case with a number of the topics, and it is therefore and it is therefore not surprising that the most coherent
topics extracted from the \textit{20 Newsgroups} corpus should occur at the settings of $K = 13$ to $K = 20$.

\begin{table}[ht]
\tiny
  \caption{Top 10 words in 6 of the topics extracted from the 20 Newsgroup text corpus where $K = 13$ for ALBU. We assign a topic category to each topic based on themes captured by most of these top ranking words. Some are mixtures of newsgroups such as \textit{comp.sys.ibm.pc.hardware} \textit{comp.sys.mac.hardware} which we have called \textit{hardware}.}
  \label{table:20news}
  \centering
  \resizebox{\textwidth}{!}{%
  \begin{tabular}{ccccccc}
    \toprule
hardware  & politics.mideast  & religion.misc  & sci.med  & sci.space  & talk.politics.guns\\
\midrule
drive  & armenian  & god  & drug  & space  & gun\\
card  & people  & one  & health  & system  & law\\
one  & said  & would  & medical  & nasa  & right\\
use  & turkish  & one  & one  & launch  & people\\
problem  & jew  & christian  & doctor  & satellite  & state\\
scsi  & one  & jesus  & use  & water  & would\\
system  & israel  & say  & disease  & may  & file\\
disk  & israeli  & think  & food  & earth  & government\\
bit  & nazi  & believe  & patient  & mission  & weapon\\
work  & woman  & bible  & day  & technology  & crime\\

    \bottomrule
  \end{tabular}}
\end{table}
Inspection of the top 10 words per topic for $K = 13$, reveals that the three algorithms yield similar results for many of the topics. The differences between Gibbs and ALBU do not seem semantically significant but for VB there are topics that could be considered incongruent mixtures of topics. For example VB extracts a topic which seems to be a mixture of \textit{talk.politics.guns} and \textit{rec.motorcycles} since it contains words like gun, firearm, weapon and handgun as well as bike motorcycle and ride. It is important to note that although the top 10 words are very similar for most of the extracted topics (especially for Gibbs and ALBU), we use a coherence window of 50 which takes more of the words into account and this is why there are differences in the coherence values, even when the top 10 words per topic are almost identical.

\begin{figure}
  \caption{Coherence scores for the 6 Newsgroups corpus. In both (a) $C_{\mathsf{V}}$ scores and (b) $C_{\mathsf{NPMI}}$ scores we see that ALBU extracts topics well at $K = 6$. }
    \begin{subfigure}{0.49\textwidth}
  \centering
  \includegraphics[width=\linewidth]{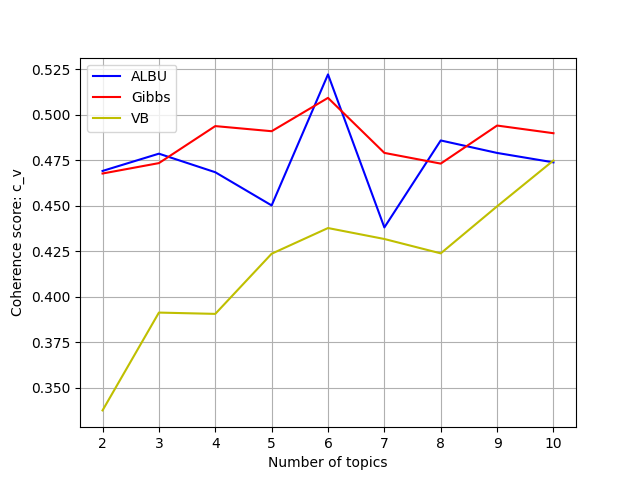}
  \caption{Gibbs shows on average higher $C_{\mathsf{V}}$ scores than the other two algorithms except for at $K = 6$. VB does significantly worse that the other two except for at  $K = 10$.\label{ART20newsSmallCV}}
  \end{subfigure}
\hfill
  \begin{subfigure}{0.49\textwidth}
  \centering
  \includegraphics[width=\linewidth]{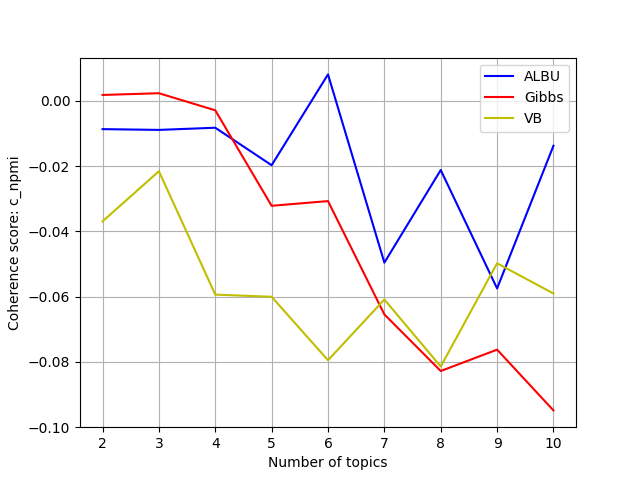}
  \caption{ALBU shows on average the highest $C_{\mathsf{NPMI}}$ score. Take note that the $C_{\mathsf{NPMI}}$ scores are below 0 for all algorithms over the whole topic range except for ALBU at $K = 6$. \label{fig:ART20newsSmalnpmi}}
  \end{subfigure}
\end{figure}

\begin{table}[ht]
\tiny
  \caption{Top 10 words in the 5  best topics extracted from the 6 Newsgroup text corpus where $K = 6$ for ALBU. We assign a topic category to each topic based on themes captured by most of these top ranking words.}
  \label{table:6news1}
  \centering
  \resizebox{\textwidth}{!}{%
  \begin{tabular}{ccccccc}
    \toprule

comp.graphics  & politics.mideast  & religion.misc  & sci.space  & 'rec.motorcycles' / stop words\\
\midrule
file  & armenian    & one       & space  & would\\
image  & jew        & people   & nasa   & bike\\
bit  & people       & god       & year  & like\\
one  & turkish    & would       & shuttle  & one\\
use  & israel       & know    & group   & people\\
also  & one         & say       & list   & get\\
system  & would     & make    & post    & think\\
program  & israeli  & good      &  news  & could\\
exercise  & case    & think  & sci       & time\\
help        & year  & bible   & launch  & food\\

    \bottomrule
  \end{tabular}}
\end{table}

\begin{table}[ht]
\tiny
  \caption{Top 10 words in the 5  best topics extracted from the 6 Newsgroup text corpus where $K = 6$ for VB. We assign a topic category to each topic based on themes captured by most of these top ranking words. We can see that two of the topics both capture information from the \textit{politics.mideast} newsgroup. The same applies for \textit{comp.graphics}. Also we can see that there are some words that do not belong in the topic (shown in bold).}
  \label{table:6news2}
  \centering
  \resizebox{\textwidth}{!}{%
  \begin{tabular}{ccccccc}
    \toprule

comp.graphics  & politics.mideast  & religion.misc  & comp.graphics/space  & politics.mideast/sci.med\\
\midrule
file  & armenian    & jew       & \textbf{space}  & bullock\\
image  & people        & jesus   & image   & israel\\
bit  & one       & one       & data  & israeli\\
one  & would    & bible       & file  & one\\
use  & turkish       & god    & available   & police\\
also  & year         & say       & system   & said\\
system  & muslim     & know    & program    & would\\
program  & armenia  & would      &  ftp  & \textbf{candida}\\
exercise  & could    & christian  & also       & time\\
help        & azerbaijan  & case   & edu  & san\\

    \bottomrule
  \end{tabular}}
\end{table}

For the \textit{6 Newsgroup} corpus the coherence values are significantly lower (since this is a significantly harder problem with only $50$ documents per newsgroup). ALBU and Gibbs again perform significantly better than does VB based on coherence measures. This agrees with actual topics extracted and shown in table  Table \ref{table:6news1} and \ref{table:6news2}. We can see that for ALBU each topic can be mostly matched to a newsgroup topic (shown in the heading of each column) where VB's topics can contain words relating to multiple newsgroup topics. For example the word with the highest probability in the topic we call \textit{comp.graphics/space} is space while the remaining 9 words all support the computer graphics topic. Also we see two of the topics are duplicated (\textit{comp.graphics} and \textit{politics.mideast}).

\subsection{Simulated corpus results}

\begin{figure}[ht]
  
    \begin{subfigure}{0.32\textwidth}
  \centering
  \includegraphics[width=\linewidth]{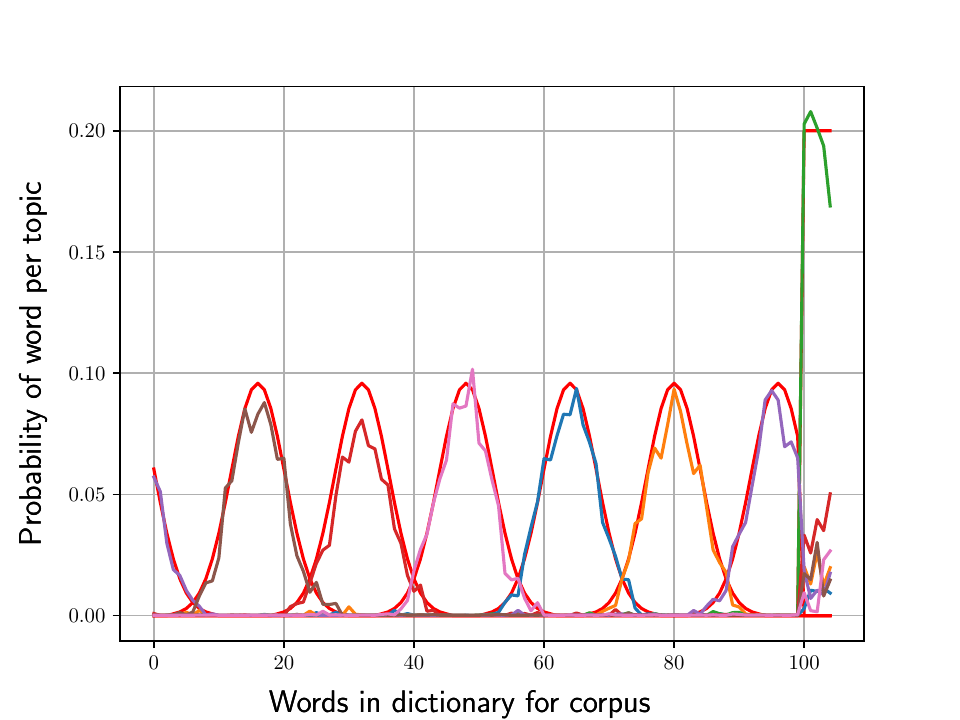}
  \caption{ALBU on the smaller simulated data set with an average KLD of  $0.11$.  }
  \end{subfigure}
    \hfill
  \begin{subfigure}{0.32\textwidth}
  \centering
  \includegraphics[width=\linewidth]{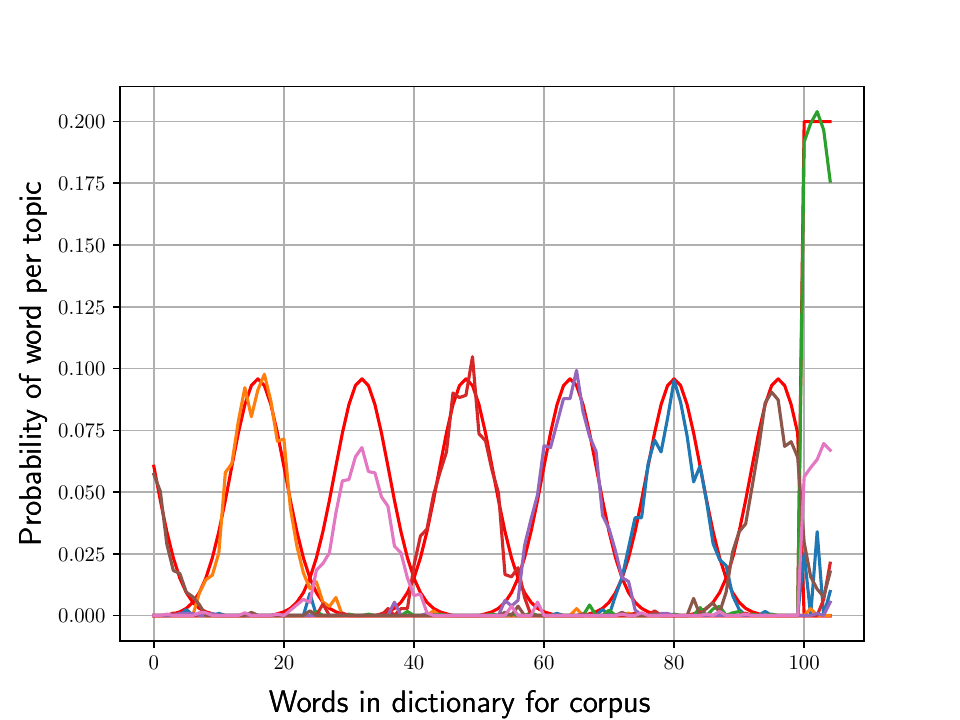}
  \caption{Collapsed Gibbs sampling on the smaller simulated data set with an average KLD of  $0.14$. }
  \end{subfigure}
  \hfill
  \begin{subfigure}{0.32\textwidth}
  \centering
  \includegraphics[width=\linewidth]{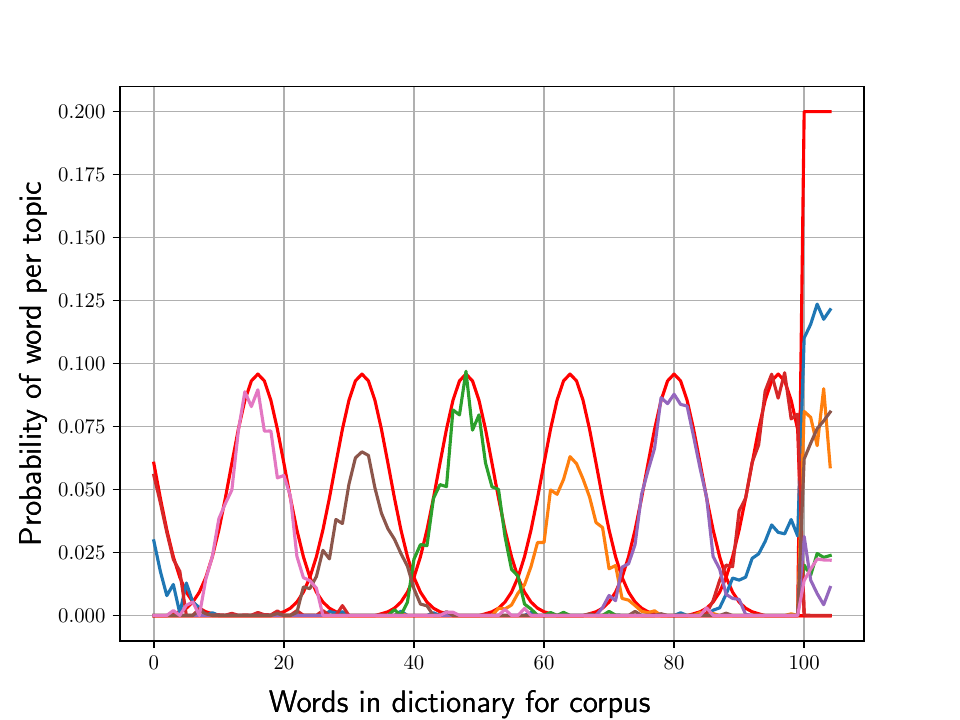}
  \caption{VB on the smaller simulated data set with an average KLD of  $0.3$.}
  \end{subfigure}
  \caption{Plot comparing true topics (shown in red), with approximate topics (coloured) for (a) ALBU and (b) Gibbs and (c) VB on the smaller data set (100 documents).\label{fig:smaller100}}
\end{figure}

We first inspect the results visually for a single corpus and show the learnt distributions in Figure \ref{fig:smaller100}. By looking at the second topic we can see that Gibbs learns this topic slightly worse than ALBU and this is reflected in the higher average KLD ($0.11$ compared to $0.14$). VB makes mistakes with several of the topics, resulting in a KLD of $0.3$. 

\begin{figure}

    \begin{subfigure}{0.49\textwidth}
  \centering
  \includegraphics[width=\linewidth]{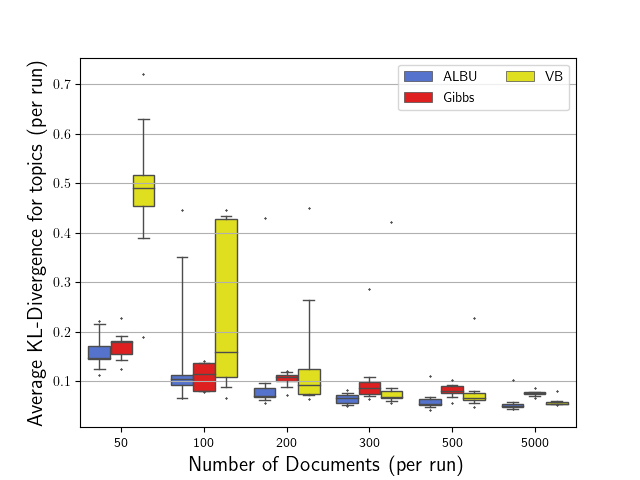}
  \caption{ALBU shows best performance over all numbers of documents. For under 200 documents, VB performs very poorly but from 300 documents it performs almost as well as ALBU adn overtakes Gibbs in performance.\label{fig:ARTsmallSimlda}}
  \end{subfigure}
\hfill
  \begin{subfigure}{0.46\textwidth}
  \centering
  \includegraphics[width=\linewidth]{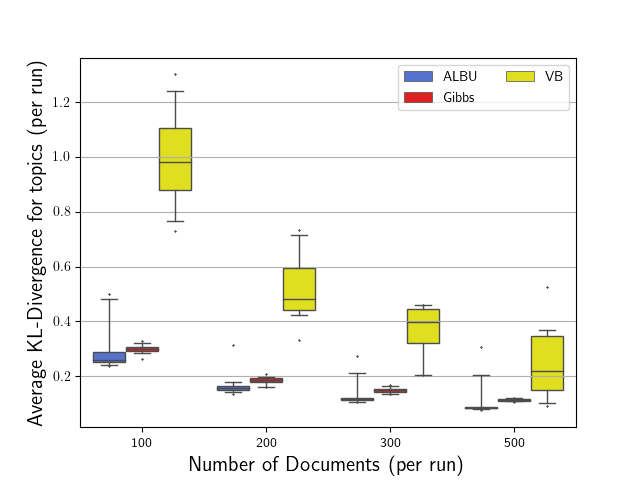}
  \caption{ALBU shows best performance over all numbers of documents closely followed by Gibbs for this bigger, more difficult, data set. VB struggles to learn the distributions well over all number of documents.\label{fig:ARTbiggerSimlda}}
  \end{subfigure}
    \caption{Boxplot showing the average KLD values for the three algorithms as the number of documents per run settings increase for (a) the small and (b) the bigger data sets.  KLD is computed over all topics for 20 runs.\label{fig:ARTSimlda}}
\end{figure}

If we compare this with Figure \ref{fig:ARTsmallSimlda} at 100 documents, we can see that these results are typical for all the corpora in this category, but do vary somewhat especially for VB.

\begin{figure}
  
    \begin{subfigure}{0.32\textwidth}
  \includegraphics[width=\linewidth]{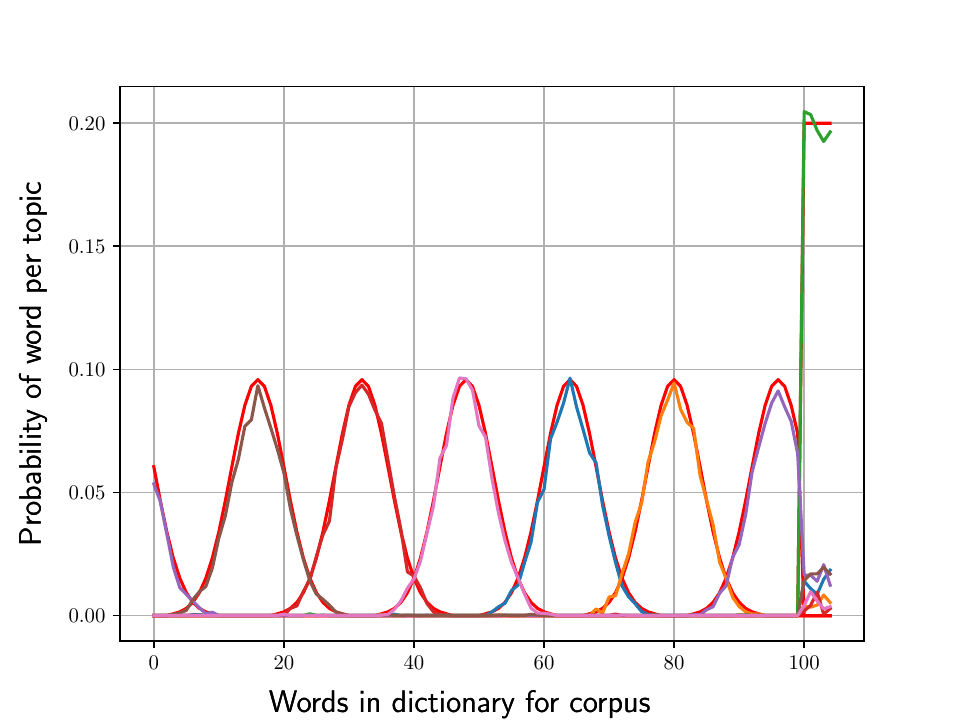}
  \caption{ALBU on the smaller simulated data set with an average KLD of $0.05$. Note how low the probabilities of the words in other optics are for the 7th topic.}
  \end{subfigure}
    \hfill
  \begin{subfigure}{0.32\textwidth}
  \includegraphics[width=\linewidth]{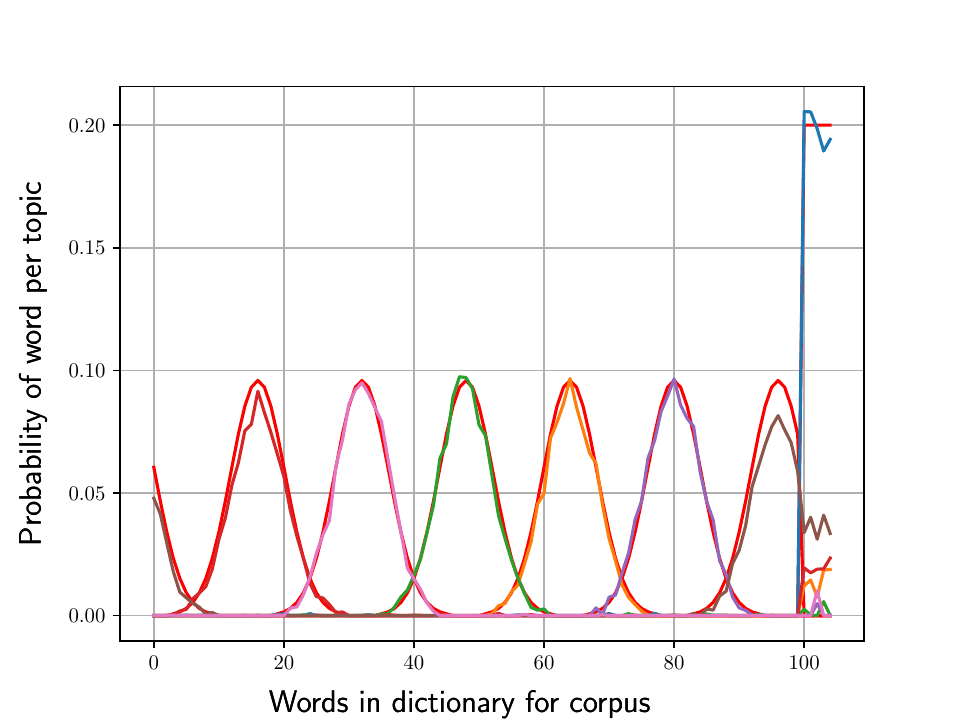}
  \caption{Collapsed Gibbs sampling on the smaller simulated data set with an average KLD of $0.07$. Performance good except for topic 5 (brown).}
  \end{subfigure}
  \hfill
  \begin{subfigure}{0.32\textwidth}
  \includegraphics[width=\linewidth]{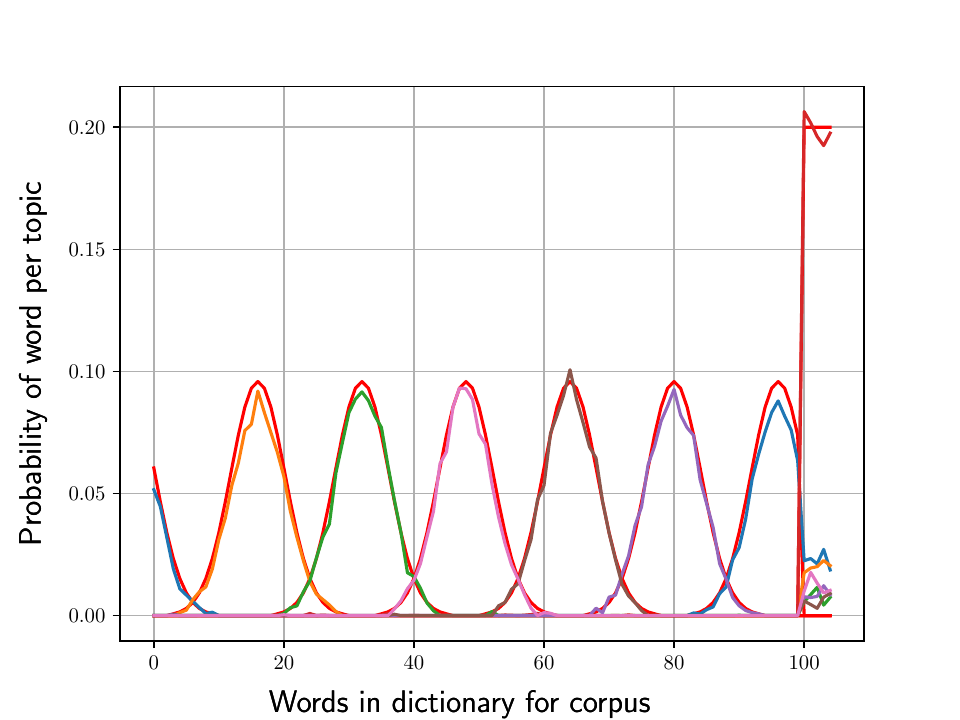}
  \caption{VB on the smaller simulated data set with an average KLD of $0.07$. Performance for topic 5 better than (b) but marginally worse than (a). }
  \end{subfigure}
  \caption{Plot comparing true topics (shown in red), with approximate topics (coloured) for (a) ALBU and (b) Gibbs and (c) VB on the smaller data set for 500 documents.\label{fig:smaller500}}
\end{figure}

Looking now one of the corpora containing 500 documents (Figure \ref{fig:smaller500}) we see that for Gibbs the topics have been learnt reasonably well except for the 6th topic which seems to have been assigned a lower weight for many of the most probably words, this weight is then assigned to words belonging in the 7th topic. The performance of VB for this corpus is very similar to ALBU (a) but topic 6 shows marginally worse performance than ALBU but better than Gibbs (b). This higlights that when we have sufficient data (on an easy enough problem) the inference method becomes less important. This is confirmed by looking again at \ref{fig:ARTsmallSimlda} but for 500 documents, we see that ALBU is clearly the best performer followed closely by VB and then Gibbs. The inferior performance by Gibbs in this case could mean that the algorithm needed even longer chains in this case to converge sufficiently each time.

In summary, looking at \ref{fig:ARTsmallSimlda}, we see that for few documents VB struggles to learn the distributions but as the number of topics increases it manages to outperform Gibbs (given the chain length) but not ALBU. This can be seen by looking at the distributions themselves and by looking at average KLD values over multiple iterations.

\begin{figure}
    \begin{subfigure}{0.32\textwidth}
  \centering
  \includegraphics[width=\linewidth]{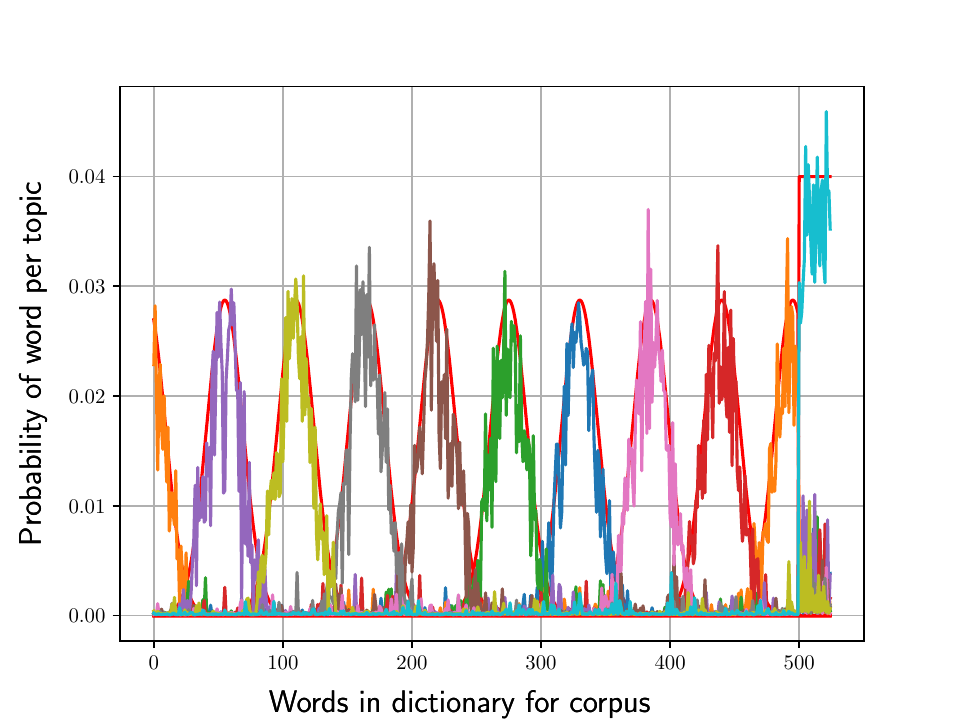}
  \caption{ALBU on the smaller simulated data set with an average KLD of $0.24$. Note how well the 10th topic is learnt.}
  \end{subfigure}
    \hfill
  \begin{subfigure}{0.32\textwidth}
  \centering
  \includegraphics[width=\linewidth]{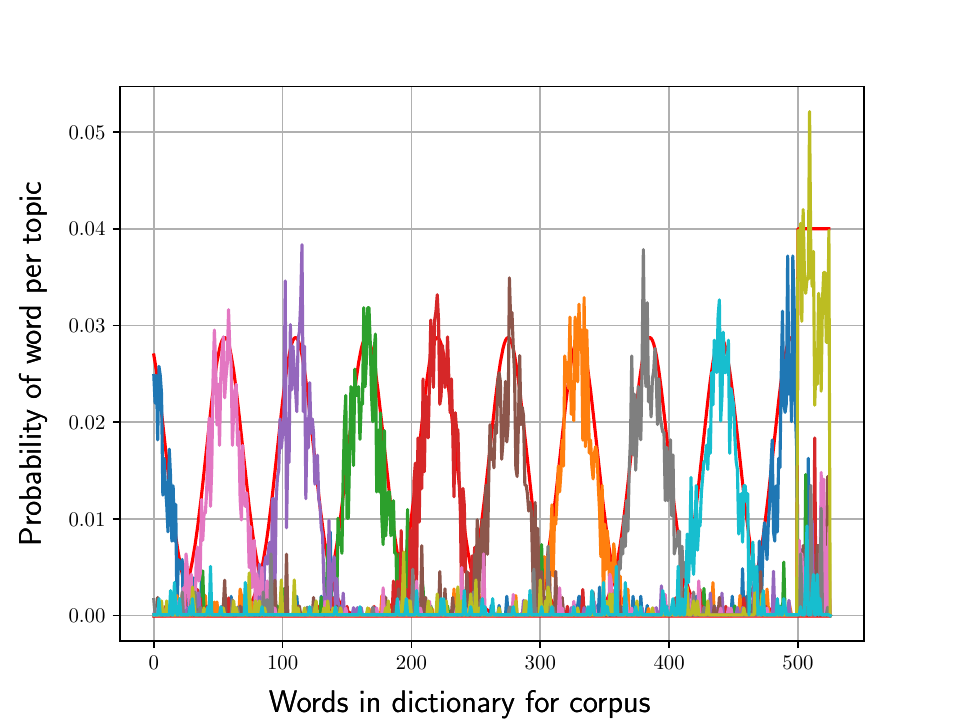}
  \caption{Collapsed Gibbs sampling on the smaller simulated data set with an average KLD of $0.32$. Performance good except for 10th topic.}
  \end{subfigure}
  \hfill
  \begin{subfigure}{0.32\textwidth}
  \centering
  \includegraphics[width=\linewidth]{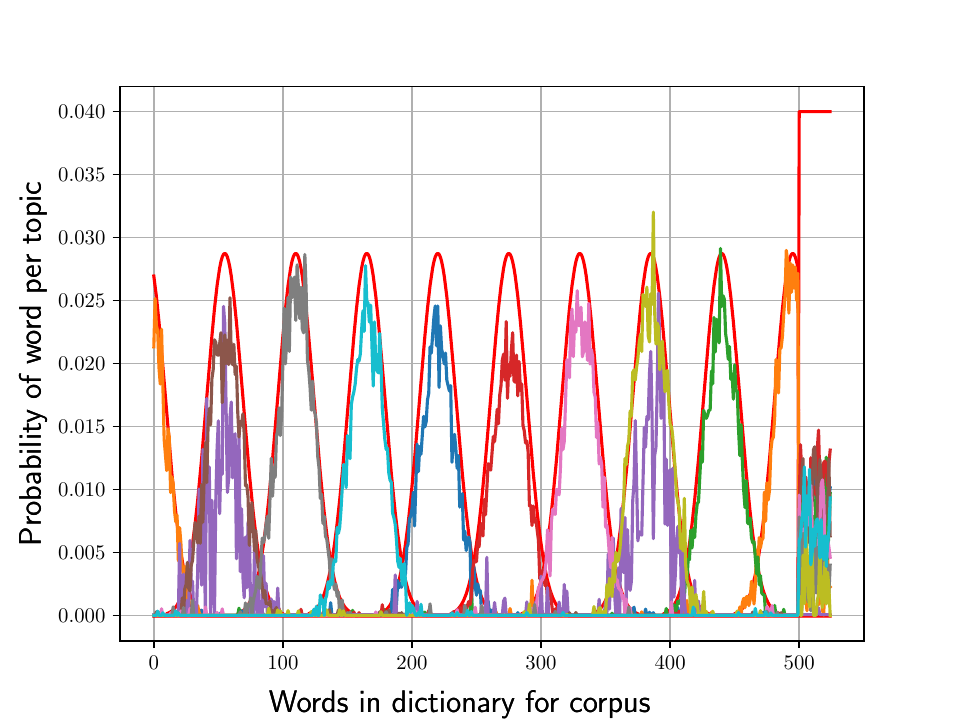}
  \caption{VB with an average KLD of $0.34$. The 9th and 10th topics missing and topics 2 and 7 are captured by two topics each. }
  \end{subfigure}
    \caption{Plot comparing true topics (shown in red), with approximate topics (coloured) for (a) ALBU and (b) Gibbs and (c) VB on the smaller data set for 100 documents.\label{fig:bigger100}}
\end{figure}

\begin{figure}
    \begin{subfigure}{0.32\textwidth}
  \includegraphics[width=\linewidth]{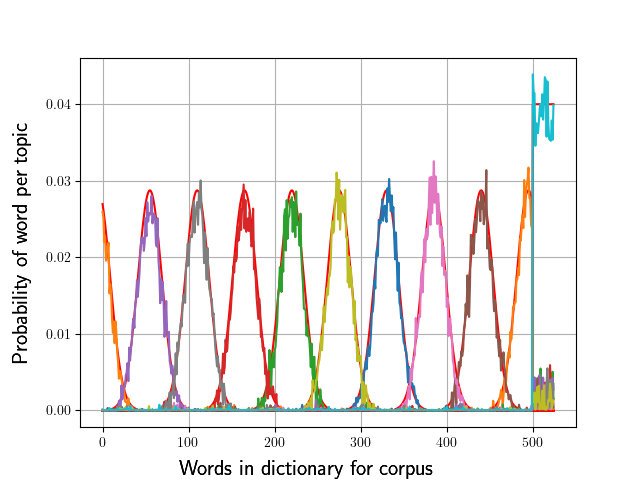}
  \caption{ALBU on the smaller simulated data set with an average KLD of $0.08$. Note how low the probabilities of the words in other optics are for the 10th topic.}
  \end{subfigure}
    \hfill
  \begin{subfigure}{0.32\textwidth}
  \includegraphics[width=\linewidth]{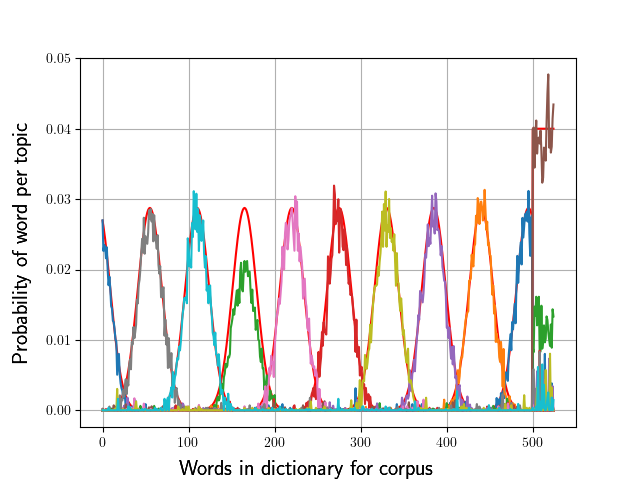}
  \caption{Collapsed Gibbs sampling on the smaller simulated data set with an average KLD of $0.12$. Performance good except for the 3rd topic (green).}
  \end{subfigure}
  \hfill
  \begin{subfigure}{0.32\textwidth}
  \includegraphics[width=\linewidth]{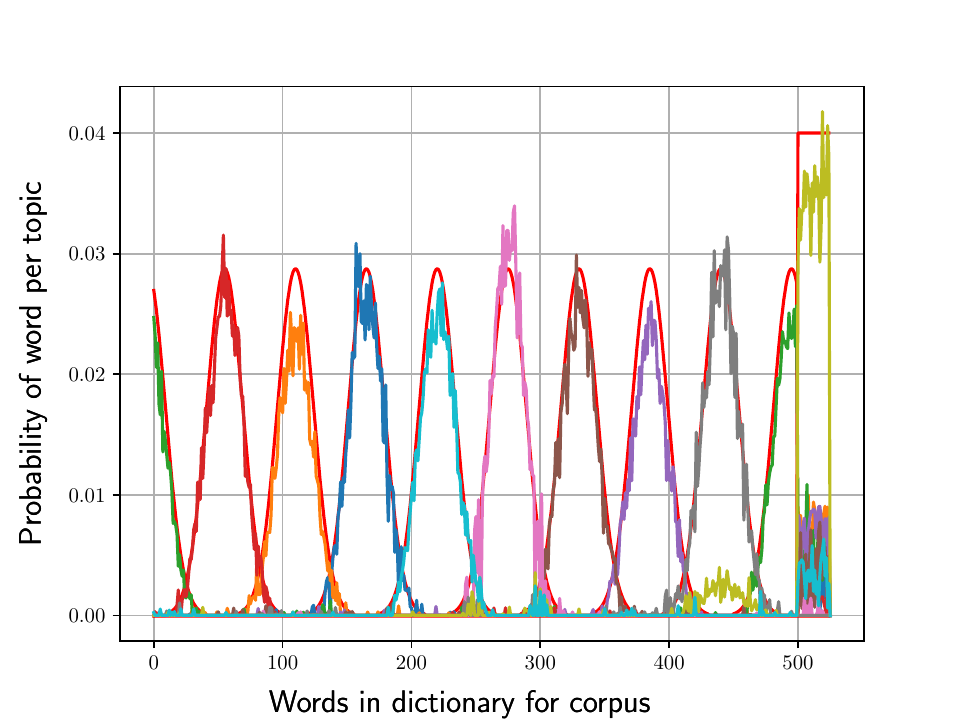}
  \caption{VB on the smaller simulated data set with an average KLD of $0.18$. Performance good except for 2nd (orange) and 10th (yellow) topics. }
  \end{subfigure}
    \caption{Plot comparing true topics (shown in red), with approximate topics (coloured) for (a) ALBU and (b) Gibbs and (c) VB on the bigger data set for 500 documents.\label{fig:bigger500}}
\end{figure}

For the bigger data set we inspect the results of one of the corpora containing 100 documents in Figure \ref{fig:bigger100} and one containing 500 documents Figure \ref{fig:bigger500}. We can see that in both figures ALBU performs best and has lowest KLD values. Gibbs performs more similarly to ALBU than VB for 100 documents but for 500 documents still does not completely learn the distributions correctly (as well as ALBU does). VB struggles to learn the distributions correctly for 100 documents but by 500 manages to learn the distributions mostly correctly (almost as well as the other two algorithms). Note that in Figure \ref{fig:bigger500} some topics are assigned weight in places where only other topics exist (note the yellow topic that should only have weight in the place of topic 10 but also has some weight in topic 8). 

In Figure \ref{fig:ARTbiggerSimlda} we see the results for the bigger data set for all corpora. It is evident that VB struggles to learn the distribution especially for corpora with fewer documents. Gibbs and ALBU perform similarly but it seems that ALBU is more likely to learn the distributions correctly. This is a harder problem to solve than the smaller data set since there are many more topics per document which means that there is more topic overlap. By taking both data sets into account it is clear that ALBU is more consistent in learning the true distributions.

\section{Discussion and Conclusion}
 Variational Bayes, the standard approximate inference algorithm for LDA, performs well at aspect identification when applied to large corpora of documents but when applied to smaller domain-specific data sets, performance can be poor. In this article we present an alternative approximate technique, based on Loopy Belief update (LBU) for LDA that is more suited to small or difficult data sets because more information is retained (we do not use expectations but rather full distributions).
 
We test our ALBU implementation against the Python Gensim VB implementation on a number of difficult data sets. We benchmark these results by including the standard collapsed Gibbs sampling algorithm (allowing an extended period for convergence). The performance for ALBU is significantly better than that of VB and comparable to that of Gibbs but at a lower computational cost. We further explore the performance of these algorithms on simulated corpora where we clearly see ALBU's superior performance. 

Our results confirm that ALBU shows similar performance to collapsed Gibbs sampling but with more consistent results (we do not need to wait as long for convergence). The aim of this article, however, is to compare inference methods on the full LDA graphical model (not the collapsed version of the model). We suspect that the performance of ALBU is very similar to that of Collapsed Variational Bayes 0 order approximation (CVBO) \cite{asuncion2009smoothing} and this could present interesting further work.

Since VB is the algorithmic equivalent of VMP, we can compare ALBU to VMP as they are both message passing algorithms. There are three main differences between the VMP and ALBU approaches. Firstly, VMP does not cancel out reverse-direction messages \cite{winn2004variational} secondly, we use full distribution where possible and do not explicitly use expected values; finally, our approximation to and from the conditional word-topic Dirichlet distributions differs form the VMP approximation (even if expected values are not used in the VMP message update). 

We believe that the combination of these three effects is the reason why ALBU outperforms VB and that the main performance enhancement comes from using full distributions and not conditional ones (where only expectations are passed on as messages). An added advantage is that our updates are simple and do not require the use of Digamma functions.

The findings in this article indicate that although we live in a word where large volumes of data are available for model training and inference, using fully Bayesian principles can provide improved performance when problems get hard (without significant extra complexity). Based on our findings we invite others to apply the principles that we have used in ALBU to other similar graphical models with Dirichlet-Categorical relationships.


\bibliographystyle{unsrtnat}


\begin{thebibliography}{48}
\providecommand{\natexlab}[1]{#1}
\providecommand{\url}[1]{\texttt{#1}}
\expandafter\ifx\csname urlstyle\endcsname\relax
  \providecommand{\doi}[1]{doi: #1}\else
  \providecommand{\doi}{doi: \begingroup \urlstyle{rm}\Url}\fi

\bibitem[Blei et~al.(2003)Blei, Ng, and Jordan]{blei2003latent}
David~M Blei, Andrew~Y Ng, and Michael~I Jordan.
\newblock Latent dirichlet allocation.
\newblock \emph{Journal of machine Learning research}, 3\penalty0
  (Jan):\penalty0 993--1022, 2003.

\bibitem[Buntine(2002)]{buntine2002variational}
Wray Buntine.
\newblock Variational extensions to em and multinomial pca.
\newblock In \emph{European Conference on Machine Learning}, pages 23--34.
  Springer, 2002.

\bibitem[Backenroth et~al.(2017)Backenroth, He, Kiryluk, Boeva, Pethukova,
  Khurana, Christiano, Buxbaum, and Ionita-Laza]{backenroth2017fun}
Daniel Backenroth, Zihuai He, Krzysztof Kiryluk, Valentina Boeva, Lynn
  Pethukova, Ekta Khurana, Angela Christiano, Joseph Buxbaum, and Iuliana
  Ionita-Laza.
\newblock Fun-lda: A latent dirichlet allocation model for predicting
  tissue-specific functional effects of noncoding variation.
\newblock \emph{bioRxiv}, page 069229, 2017.

\bibitem[Pritchard et~al.(2000)Pritchard, Stephens, and
  Donnelly]{pritchard2000inference}
Jonathan~K Pritchard, Matthew Stephens, and Peter Donnelly.
\newblock Inference of population structure using multilocus genotype data.
\newblock \emph{Genetics}, 155\penalty0 (2):\penalty0 945--959, 2000.

\bibitem[Fei-Fei and Perona(2005)]{fei2005bayesian}
Li~Fei-Fei and Pietro Perona.
\newblock A bayesian hierarchical model for learning natural scene categories.
\newblock In \emph{2005 IEEE Computer Society Conference on Computer Vision and
  Pattern Recognition (CVPR'05)}, volume~2, pages 524--531. IEEE, 2005.

\bibitem[Sivic et~al.(2005)Sivic, Russell, Efros, Zisserman, and
  Freeman]{sivic2005discovering}
Josef Sivic, Bryan~C Russell, Alexei~A Efros, Andrew Zisserman, and William~T
  Freeman.
\newblock Discovering object categories in image collections.
\newblock 2005.

\bibitem[Wang and Grimson(2008)]{wang2008spatial}
Xiaogang Wang and Eric Grimson.
\newblock Spatial latent dirichlet allocation.
\newblock In \emph{Advances in neural information processing systems}, pages
  1577--1584, 2008.

\bibitem[Feuerriegel and Pr{\"o}llochs(2018)]{feuerriegel2018investor}
Stefan Feuerriegel and Nicolas Pr{\"o}llochs.
\newblock Investor reaction to financial disclosures across topics: An
  application of latent dirichlet allocation.
\newblock \emph{Decision Sciences}, 2018.

\bibitem[Shirota et~al.(2014)Shirota, Hashimoto, and
  Sakura]{shirota2014extraction}
Yukari Shirota, Takako Hashimoto, and Tamaki Sakura.
\newblock Extraction of the financial policy topics by latent dirichlet
  allocation.
\newblock In \emph{TENCON 2014-2014 IEEE Region 10 Conference}, pages 1--5.
  IEEE, 2014.

\bibitem[Woodbury and Manton(1982)]{woodbury1982new}
Max~A Woodbury and Kenneth~G Manton.
\newblock A new procedure for analysis of medical classification.
\newblock \emph{Methods of Information in Medicine}, 21\penalty0 (04):\penalty0
  210--220, 1982.

\bibitem[Minka and Lafferty(2002)]{minka2002expectation}
Thomas Minka and John Lafferty.
\newblock Expectation-propagation for the generative aspect model.
\newblock In \emph{Proceedings of the Eighteenth conference on Uncertainty in
  artificial intelligence}, pages 352--359. Morgan Kaufmann Publishers Inc.,
  2002.

\bibitem[Phan et~al.(2008)Phan, Nguyen, and Horiguchi]{phan2008learning}
Xuan-Hieu Phan, Le-Minh Nguyen, and Susumu Horiguchi.
\newblock Learning to classify short and sparse text \& web with hidden topics
  from large-scale data collections.
\newblock In \emph{Proceedings of the 17th international conference on World
  Wide Web}, pages 91--100. ACM, 2008.

\bibitem[Hong and Davison(2010)]{hong2010empirical}
Liangjie Hong and Brian~D Davison.
\newblock Empirical study of topic modeling in twitter.
\newblock In \emph{Proceedings of the first workshop on social media
  analytics}, pages 80--88. acm, 2010.

\bibitem[Ramage et~al.(2010)Ramage, Dumais, and
  Liebling]{ramage2010characterizing}
Daniel Ramage, Susan Dumais, and Dan Liebling.
\newblock Characterizing microblogs with topic models.
\newblock In \emph{Fourth international AAAI conference on weblogs and social
  media}, 2010.

\bibitem[Jin et~al.(2011)Jin, Liu, Zhao, Yu, and Yang]{jin2011transferring}
Ou~Jin, Nathan~N Liu, Kai Zhao, Yong Yu, and Qiang Yang.
\newblock Transferring topical knowledge from auxiliary long texts for short
  text clustering.
\newblock In \emph{Proceedings of the 20th ACM international conference on
  Information and knowledge management}, pages 775--784. ACM, 2011.

\bibitem[Weng et~al.(2010)Weng, Lim, Jiang, and He]{weng2010twitterrank}
Jianshu Weng, Ee-Peng Lim, Jing Jiang, and Qi~He.
\newblock Twitterrank: finding topic-sensitive influential twitterers.
\newblock In \emph{Proceedings of the third ACM international conference on Web
  search and data mining}, pages 261--270. ACM, 2010.

\bibitem[Basave et~al.(2014)Basave, He, and Xu]{basave2014automatic}
Amparo Elizabeth~Cano Basave, Yulan He, and Ruifeng Xu.
\newblock Automatic labelling of topic models learned from twitter by
  summarisation.
\newblock In \emph{Proceedings of the 52nd Annual Meeting of the Association
  for Computational Linguistics (Volume 2: Short Papers)}, pages 618--624,
  2014.

\bibitem[Nugroho et~al.(2015)Nugroho, Molla-Aliod, Yang, Zhong, Paris, and
  Nepal]{nugroho2015incorporating}
Robertus Nugroho, Diego Molla-Aliod, Jian Yang, Youliang Zhong, Cecile Paris,
  and Surya Nepal.
\newblock Incorporating tweet relationships into topic derivation.
\newblock In \emph{Conference of the Pacific Association for Computational
  Linguistics}, pages 177--190. Springer, 2015.

\bibitem[Mehrotra et~al.(2013)Mehrotra, Sanner, Buntine, and
  Xie]{mehrotra2013improving}
Rishabh Mehrotra, Scott Sanner, Wray Buntine, and Lexing Xie.
\newblock Improving lda topic models for microblogs via tweet pooling and
  automatic labeling.
\newblock In \emph{Proceedings of the 36th international ACM SIGIR conference
  on Research and development in information retrieval}, pages 889--892. ACM,
  2013.

\bibitem[Yan et~al.(2013)Yan, Guo, Lan, and Cheng]{yan2013biterm}
Xiaohui Yan, Jiafeng Guo, Yanyan Lan, and Xueqi Cheng.
\newblock A biterm topic model for short texts.
\newblock In \emph{Proceedings of the 22nd international conference on World
  Wide Web}, pages 1445--1456. ACM, 2013.

\bibitem[Albakour et~al.(2013)Albakour, Macdonald, Ounis,
  et~al.]{albakour2013sparsity}
M~Albakour, Craig Macdonald, Iadh Ounis, et~al.
\newblock On sparsity and drift for effective real-time filtering in
  microblogs.
\newblock In \emph{Proceedings of the 22nd ACM international conference on
  Information \& Knowledge Management}, pages 419--428. ACM, 2013.

\bibitem[Jabeur et~al.(2012)Jabeur, Tamine, and Boughanem]{jabeur2012uprising}
Lamjed~Ben Jabeur, Lynda Tamine, and Mohand Boughanem.
\newblock Uprising microblogs: A bayesian network retrieval model for tweet
  search.
\newblock In \emph{Proceedings of the 27th annual ACM symposium on applied
  computing}, pages 943--948. ACM, 2012.

\bibitem[Celikyilmaz et~al.(2010)Celikyilmaz, Hakkani-T{\"u}r, and
  Feng]{celikyilmaz2010probabilistic}
Asli Celikyilmaz, Dilek Hakkani-T{\"u}r, and Junlan Feng.
\newblock Probabilistic model-based sentiment analysis of twitter messages.
\newblock In \emph{2010 IEEE Spoken Language Technology Workshop}, pages
  79--84. IEEE, 2010.

\bibitem[Quan et~al.(2015)Quan, Kit, Ge, and Pan]{quan2015short}
Xiaojun Quan, Chunyu Kit, Yong Ge, and Sinno~Jialin Pan.
\newblock Short and sparse text topic modeling via self-aggregation.
\newblock In \emph{Twenty-Fourth International Joint Conference on Artificial
  Intelligence}, 2015.

\bibitem[Sokolova et~al.(2016)Sokolova, Huang, Matwin, Ramisch, Sazonova,
  Black, Orwa, Ochieng, and Sambuli]{sokolova2016topic}
Marina Sokolova, Kanyi Huang, Stan Matwin, Joshua Ramisch, Vera Sazonova, Renee
  Black, Chris Orwa, Sidney Ochieng, and Nanjira Sambuli.
\newblock Topic modelling and event identification from twitter textual data.
\newblock \emph{arXiv preprint arXiv:1608.02519}, 2016.

\bibitem[Wang and McCallum(2006)]{wang2006topics}
Xuerui Wang and Andrew McCallum.
\newblock Topics over time: a non-markov continuous-time model of topical
  trends.
\newblock In \emph{Proceedings of the 12th ACM SIGKDD international conference
  on Knowledge discovery and data mining}, pages 424--433. ACM, 2006.

\bibitem[Rosen-Zvi et~al.(2004)Rosen-Zvi, Griffiths, Steyvers, and
  Smyth]{rosen2004author}
Michal Rosen-Zvi, Thomas Griffiths, Mark Steyvers, and Padhraic Smyth.
\newblock The author-topic model for authors and documents.
\newblock In \emph{Proceedings of the 20th conference on Uncertainty in
  artificial intelligence}, pages 487--494. AUAI Press, 2004.

\bibitem[Winn(2004)]{winn2004variational}
John~Michael Winn.
\newblock \emph{Variational message passing and its applications}.
\newblock PhD thesis, Citeseer, 2004.

\bibitem[Blei et~al.(2017)Blei, Kucukelbir, and McAuliffe]{blei2017variational}
David~M Blei, Alp Kucukelbir, and Jon~D McAuliffe.
\newblock Variational inference: A review for statisticians.
\newblock \emph{Journal of the American Statistical Association}, 112\penalty0
  (518):\penalty0 859--877, 2017.

\bibitem[Attias(2000)]{attias2000variational}
Hagai Attias.
\newblock A variational baysian framework for graphical models.
\newblock In \emph{Advances in neural information processing systems}, pages
  209--215, 2000.

\bibitem[Winn and Bishop(2005)]{winn2005variational}
John Winn and Christopher~M Bishop.
\newblock Variational message passing.
\newblock \emph{Journal of Machine Learning Research}, 6\penalty0
  (Apr):\penalty0 661--694, 2005.

\bibitem[Foulds et~al.(2013)Foulds, Boyles, DuBois, Smyth, and
  Welling]{foulds2013stochastic}
James Foulds, Levi Boyles, Christopher DuBois, Padhraic Smyth, and Max Welling.
\newblock Stochastic collapsed variational bayesian inference for latent
  dirichlet allocation.
\newblock In \emph{Proceedings of the 19th ACM SIGKDD international conference
  on Knowledge discovery and data mining}, pages 446--454. ACM, 2013.

\bibitem[Rehurek and Sojka(2010)]{rehurek2010software}
Radim Rehurek and Petr Sojka.
\newblock Software framework for topic modelling with large corpora.
\newblock In \emph{In Proceedings of the LREC 2010 Workshop on New Challenges
  for NLP Frameworks}. Citeseer, 2010.

\bibitem[Pedregosa et~al.(2011)Pedregosa, Varoquaux, Gramfort, Michel, Thirion,
  Grisel, Blondel, Prettenhofer, Weiss, Dubourg, Vanderplas, Passos,
  Cournapeau, Brucher, Perrot, and Duchesnay]{pedregosa2011scikit}
F.~Pedregosa, G.~Varoquaux, A.~Gramfort, V.~Michel, B.~Thirion, O.~Grisel,
  M.~Blondel, P.~Prettenhofer, R.~Weiss, V.~Dubourg, J.~Vanderplas, A.~Passos,
  D.~Cournapeau, M.~Brucher, M.~Perrot, and E.~Duchesnay.
\newblock Scikit-learn: Machine learning in {P}ython.
\newblock \emph{Journal of Machine Learning Research}, 12:\penalty0 2825--2830,
  2011.

\bibitem[Koller et~al.(2009)Koller, Friedman, and
  Bach]{koller2009probabilistic}
Daphne Koller, Nir Friedman, and Francis Bach.
\newblock \emph{Probabilistic graphical models: principles and techniques}.
\newblock MIT press, 2009.

\bibitem[Lauritzen and Spiegelhalter(1988)]{lauritzen1988local}
Steffen~L Lauritzen and David~J Spiegelhalter.
\newblock Local computations with probabilities on graphical structures and
  their application to expert systems.
\newblock \emph{Journal of the Royal Statistical Society: Series B
  (Methodological)}, 50\penalty0 (2):\penalty0 157--194, 1988.

\bibitem[Heskes(2003)]{heskes2003stable}
Tom Heskes.
\newblock Stable fixed points of loopy belief propagation are local minima of
  the bethe free energy.
\newblock In \emph{Advances in neural information processing systems}, pages
  359--366, 2003.

\bibitem[Zeng(2012)]{zeng2012topic}
Jia Zeng.
\newblock A topic modeling toolbox using belief propagation.
\newblock \emph{Journal of Machine Learning Research}, 13\penalty0
  (Jul):\penalty0 2233--2236, 2012.

\bibitem[Zeng et~al.(2012)Zeng, Cheung, and Liu]{zeng2012learning}
Jia Zeng, William~K Cheung, and Jiming Liu.
\newblock Learning topic models by belief propagation.
\newblock \emph{IEEE Transactions on Pattern Analysis and Machine
  Intelligence}, 35\penalty0 (5):\penalty0 1121--1134, 2012.

\bibitem[Teh et~al.(2007)Teh, Newman, and Welling]{teh2007collapsed}
Yee~W Teh, David Newman, and Max Welling.
\newblock A collapsed variational bayesian inference algorithm for latent
  dirichlet allocation.
\newblock In \emph{Advances in neural information processing systems}, pages
  1353--1360, 2007.

\bibitem[Asuncion et~al.(2009)Asuncion, Welling, Smyth, and
  Teh]{asuncion2009smoothing}
Arthur Asuncion, Max Welling, Padhraic Smyth, and Yee~Whye Teh.
\newblock On smoothing and inference for topic models.
\newblock In \emph{Proceedings of the twenty-fifth conference on uncertainty in
  artificial intelligence}, pages 27--34. AUAI Press, 2009.

\bibitem[pyp()]{pypi}
Python package index - pypi.
\newblock URL \url{https://pypi.org/}.

\bibitem[Mcauliffe and Blei(2008)]{mcauliffe2008supervised}
Jon~D Mcauliffe and David~M Blei.
\newblock Supervised topic models.
\newblock In \emph{Advances in neural information processing systems}, pages
  121--128, 2008.

\bibitem[Chang et~al.(2009)Chang, Gerrish, Wang, Boyd-Graber, and
  Blei]{chang2009reading}
Jonathan Chang, Sean Gerrish, Chong Wang, Jordan~L Boyd-Graber, and David~M
  Blei.
\newblock Reading tea leaves: How humans interpret topic models.
\newblock In \emph{Advances in neural information processing systems}, pages
  288--296, 2009.

\bibitem[R{\"o}der et~al.(2015)R{\"o}der, Both, and
  Hinneburg]{roder2015exploring}
Michael R{\"o}der, Andreas Both, and Alexander Hinneburg.
\newblock Exploring the space of topic coherence measures.
\newblock In \emph{Proceedings of the eighth ACM international conference on
  Web search and data mining}, pages 399--408. ACM, 2015.

\bibitem[Wallach et~al.(2009)Wallach, Mimno, and
  McCallum]{wallach2009rethinking}
Hanna~M Wallach, David~M Mimno, and Andrew McCallum.
\newblock Rethinking lda: Why priors matter.
\newblock In \emph{Advances in neural information processing systems}, pages
  1973--1981, 2009.

\bibitem[Mukherjee and Blei(2009)]{mukherjee2009relative}
Indraneel Mukherjee and David~M Blei.
\newblock Relative performance guarantees for approximate inference in latent
  dirichlet allocation.
\newblock In \emph{Advances in Neural Information Processing Systems}, pages
  1129--1136, 2009.

\bibitem[Minka(2000)]{minka2000estimating}
Thomas Minka.
\newblock Estimating a dirichlet distribution, 2000.

\end{thebibliography}

\end{document}